%% file: main.tex
\documentclass[11pt, copyright, logo]{gdm_tech_report_template/googledeepmind}
\usepackage[authoryear, sort&compress, round]{natbib}
\usepackage{bbm}
\usepackage{enumitem}
\setlist{nosep}
\usepackage{listings}
\usepackage{adjustbox}  
\usepackage{graphicx}
\usepackage{caption}
\usepackage{tabularx}
\usepackage{booktabs}
\usepackage{subcaption}
\usepackage{xcolor}
\usepackage{hyperref}
\usepackage{verbatim}
\usepackage[normalem]{ulem}
\usepackage{float}
\usepackage{cleveref}
\usepackage{wrapfig}
\usepackage{makecell}
\usepackage{multirow}
\usepackage{fancyvrb}
\usepackage[normalem]{ulem}
\usepackage{url}
\input{math_commands.tex}

\input{abbreviations.tex}  

\title{Performance Prediction for Large Systems via Text-to-Text Regression}

\author[1,2]{Yash Akhauri$^*$}
\author[1]{Bryan Lewandowski}

\author[1]{Cheng-Hsi Lin}
\author[1]{Adrian N. Reyes}
\author[3]{Grant C. Forbes}
\author[1]{Arissa Wongpanich}
\author[1]{Bangding Yang}

\author[2]{Mohamed S. Abdelfattah}
\author[1]{Sagi Perel}
\author[1]{Xingyou Song}

\affil[1]{Google}
\affil[2]{Cornell University}
\affil[3]{North Carolina State University}

\begin{abstract}
In many industries, predicting metric outcomes of large systems is a fundamental problem, driven largely by traditional tabular regression. However, such methods struggle on complex systems data in the wild such as configuration files or system logs, where feature engineering is often infeasible. We propose text-to-text regression as a general, scalable alternative. For predicting resource efficiency on Borg, Google's massive compute cluster scheduling system, a 60M parameter encoder-decoder, trained from random initialization, achieves up to a near perfect 0.99 (0.9 average) rank correlation across the entire fleet, and 100x lower MSE than tabular approaches. The model also easily adapts to new tasks in only 500 few-shot examples and captures the densities of complex outcome distributions. Ablation studies highlight the importance of using encoders, increasing sequence length, and the model's inherent uncertainty quantification. These findings pave the way for universal simulators of real-world outcomes.
\end{abstract}

\begin{document}
\maketitle

\section{Introduction}
\textit{Performance prediction} has been an important problem for various industrial system use cases, ranging from latency prediction \citep{latency_prediction}, execution times \citep{ernst}, scheduling conflicts \citep{network_function}, and transaction response timing \citep{transaction_response}. While traditional methods have significantly relied on expert domain knowledge to model specific numeric metrics, more recent data-centric techniques have predominantly used machine-learning based regression methods by training models which predict metric $y$ given a set of features $x$.

\begin{figure}[h]
    \centering
    \includegraphics[width=0.98\textwidth]{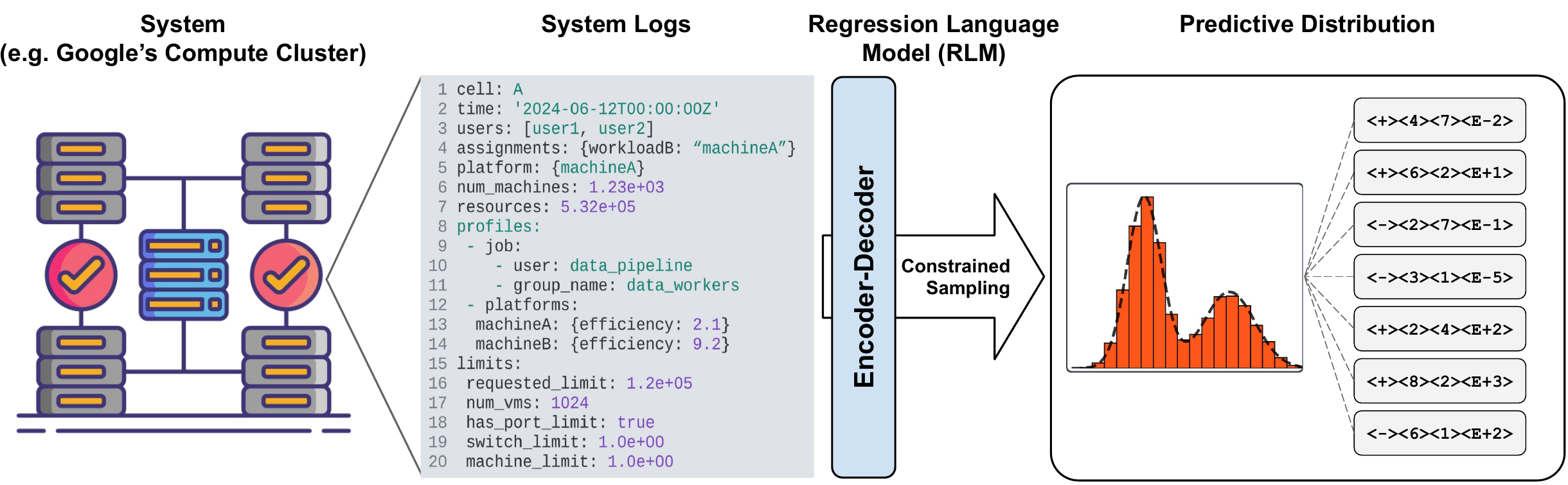}
    \caption{Overview: Using $(x,y)$ metric data collected from a variety of system logs, we train a encoder-decoder via standard next-token cross-entropy for performance prediction.}
    \label{fig:intro}
\end{figure}

However, standard machine learning regression techniques such as random forests and multi-layer perceptrons require the features $x$ to be represented as flat fixed-length tensors, which can be difficult for certain prediction problems with highly non-tabular representations. Even if such feature engineering concerns are resolved, compressing features only as numbers without much context can be highly lossy, and end up treating the system as a blackbox \citep{blackbox}, which can drastically decrease predictive performance.

Graybox techniques \citep{graybox_analytical_modeling} have attempted to mitigate these concerns by combining both domain-expertise and machine learning, by e.g. learning the coefficients to an already-derived symbolic expression (e.g. latency of a website as a linear function of users). These approaches often require large amounts of prior knowledge on the relationship between $x$ and $y$, and are very restrictive in their applicability.

However, the advent of language models has since enabled \textit{text-to-text regression}, which can avoid many of the issues regarding tensor-based feature engineering. Recently, \cite{omnipred} showed that models even of relatively small size (<1B parameters), which we term as \textit{Regression Language Models (RLMs)}, are capable of enormous transfer learning if trained over large amounts of $(x,y)$ regression data, being able to improve prediction accuracy simply by absorbing information from multiple different regression tasks. Surprisingly, training with next-token prediction over numeric representations via cross-entropy loss is also quite sample efficient \citep{song2025decodingbasedregression}, capable of matching or even outperforming standard regression methods when given the same data.

In this paper, we demonstrate RLMs can be applied to even simulating efficiency outcomes by training over large amounts of data from large-scale industrial systems such as Google's ``Borg'' compute cluster, which can drastically reduce production costs. In summary, our contributions are as follows:
\begin{itemize}
\item RLMs are capable of predicting numeric outcomes, such as efficiency metrics of industrial systems (e.g. Google's entire compute cluster), with high precision over complex feature representations and multi-modal outcome distributions.
\item By fine-tuning over very small amounts of new $(x,y)$ training data, such models are highly capable of few-shot adaptation, wherein a pretrained model is able to remain highly accurate on unseen compute clusters or scenarios.
\item Comprehensive ablations provide insights into the performance improvement effects of changing sequence length, model size, feature observability, architecture, learning rate, and early stopping, in addition to the model's natural uncertainty quantification abilities.
\end{itemize}

\section{System Performance Prediction}
\subsection{Background}
We provide a brief overview of Borg \citep{borg}, the compute management system at Google. The system comprises of a centralized manager for scheduling jobs, and a list of machines located in different \emph{cells} with compute resources. Each user may send in a \emph{task request}, containing a binary executable with additional details such as resources needed, number of replicas, and so on. The role of the manager is to place each task on an appropriate machine (i.e., one where the requisite resources are available). 

To effectively allocate computing tasks to resources, the scheduler must know the resource consumption (i.e. utilization) of a task on that machine and ultimately how much useful work can be done (i.e. productivity), in order to execute a specialized bin-packing algorithm for assigning jobs, to maximize total productivity across the entire compute cluster.


For a given process, this productivity metric is formally defined as Millions of Instructions Per Second per Google Computing Unit \citep{gcu_definition} of usage, or ``MIPS per GCU''. Intuitively, this metric represents the amount of work that gets done per unit of time by using per unit of computing resource. Empirically, \textit{after} job assignment has been made, MIPS per GCU can be computed periodically by profiling the total instructions executed and total CPU cycles consumed in a 10-second time window. 

This metric can be affected by numerous factors, ranging from current memory and CPU usage, hardware type, and workload mixes across numerous machines. Furthermore, even variations in time can even have an effect, as workloads within a physical cell depend on the usage patterns by users within the same geographic location. Additionally, hyperparameters affecting the behavior of the underlying bin-packing algorithm may also strongly affect the outcome.

Fortunately, Google has developed a digital twin of Borg, a sophisticated backtesting framework which uses checkpoint files of real-world clusters to replicate the state of those clusters, perform the scheduling algorithm, and determine the aggregate cluster-wide MIPS per GCU across all machines. However, due to the inherently sequential nature, generating even a single outcome can require between 1 to 18 hours of computation regardless of resources used.

Fortunately, such outcomes are logged as valuable offline datasets for training cheap regression models, albeit from limited data. If a regressor is able to accurately predict this metric with negligible inference time, this can lead to large savings not just from avoiding computation, but also from faster optimization of the MIPS per GCU overall. For instance, Google Vizier \citep{google_vizier} is regularly used to tune the Borg scheduler's hyperparameters, but unfortunately its Gaussian Process regressor \citep{vizier_gp} can at most only observe tabular data formats, drastically limiting its predictive and overall optimization performance.




\subsection{Prediction Task}
The goal is to predict the MIPS per GCU efficiency metric, i.e. a single observed floating point number, after a specialized bin-packing algorithm has been performed, given the initial state of the compute cluster, time window used for collecting metric data, and hyperparameters of the scheduling algorithm. More specifically, the following available information may all be used as input features:

\begin{itemize}
\item Cluster name (i.e. ``cell'').
\item Physical location of the cell.
\item The window of time used to collect the state of the cell and the exact time the performance profiles were collected.
\item Hyperparameters affecting the behavior of the Borg scheduler's bin-packing algorithm.
\item Concurrent entities or teams using the largest amount of compute within this cell.
\item Distribution and network hierarchy of hardware platforms installed in the cell.
\item A list of job-on-machine performance profiling results.
\end{itemize}

All of these features are presumed to be important based on minimal domain knowledge, but it is not apriori clear how exactly they may affect this metric. For example, the efficiency metric is heavily dependent on the cell, which is the specific cluster of machines and hardware, but the actual cell name itself is not a true feature. Features such as timing windows and hyperparameter names can be shared or correlated across different tasks, but may affect efficiency metrics differently.

Furthermore, the prediction task can be dynamic, as new features values may appear over time, due to platform upgrades and hardware changes. The features may also contain varying types, e.g. categorical hyperparameters, different subsets of hardware types, and cyclical timestamps. Additional complexity arises due to the deep nested nature of many of the features -- e.g. job-on-machine profilings containing information about jobs and which machines they utilize, which also contain information about hardware profiles, and so forth. Lastly, the metric may also possess various forms of noise, some dependent on the features observed by the regressor, as we discuss below.

\subsection{Observed Features vs. Uncertainty}
\label{subsec:epistemic_uncertainty}
The most broad abstraction to define our prediction task is to assume there is an underlying distribution $p(y|x)$ where $x$ is the \textit{full state} of the world. Thus the randomness induced by $p(y|x)$ can be considered irreducible noise, also known as \textit{aleatoric uncertainty} \citep{aleatoric_epistemic}, which limits the optimal performance of a pointwise regressor. For Borg, this can come from the inherent randomness of the bin-packing algorithm and stochastic load/demand (e.g. for user-facing services).

More formally, based on the well-known \textit{Bias-Variance decomposition}, the expected squared error of a pointwise regressor $f_{\theta}$ per input $x$ is lower bounded by the variance of the $y$:
\begin{equation}
\mathbb{E}_{y \sim p(y|x) } \left[ (y - f_{\theta}(x))^{2} \right] \ge \text{Var}(y | x)
\end{equation}
However, if a regressor is only able to observe a partial subset or limited representation $\phi(x)$ of the full state, it will be unable to distinguish separate $x, x'$ if $\phi(x) = \phi(x')$. This lack of distinguishability induces \textit{epistemic uncertainty}, and instead leads to an even higher right hand variance term $\text{Var}(y| \phi(x))$, limiting the regressor's optimal performance even more (theory in Appendix \ref{appendix:epistemic}). For practical purposes, given an offline test dataset $\mathcal{D} = \{(x_i, y_i)\}_{i \ge 0}$, we can estimate $\text{TotalVariance}_{\phi}(\mathcal{D})$ to provide lower bounds on the mean squared error (MSE) of regression methods evaluated over all test data, as:
\begin{equation}
\text{TotalVariance}_{\phi}(\mathcal{D}) = \frac{1}{K}\sum_{k=1}^{K} \text{Var}(y|x \in \mathcal{X}_k)
\end{equation}
where $\mathcal{X}_1, \ldots, \mathcal{X}_K$ are equivalence classes partitioning $\mathcal{D}$, i.e. $x, x' \in \mathcal{X}_{k}$ iff $\phi(x) = \phi(x')$, and variance is calculated empirically over the set of all $y$-values obtained from inputs within $\mathcal{X}_k$. Note that if $\phi$ is ``null'' (no observed features), this expression would simply be the variance of the entire $y$-population.

Similar bounds exist for other regression-based metrics, e.g. rank correlations, since determining the relative rankings of $y$-values from the same $x$-equivalence class would be impossible, and a density estimator $p_{\theta}$'s log-likelihood would be bounded by the conditional entropy of $p(y|\phi(x))$. In any case, it thus would be ideal if the regressor is able to maximize the amount of features it observes in order to minimize epistemic uncertainty.

\subsection{Why Text-based Regression?}
Representing all of the features above into one single fixed-length tensor in $\mathbb{R}^d$ will be notoriously difficult if using a traditional tabular regressor such as a multi-layer perceptron (MLP) or random forest. Many of the features such as the list of jobs and platforms can contain arbitrary cardinalities per $x$, and would therefore need to be grouped using hand-selected methods or heuristics. Even if a useful set of hand-selected metrics can be found, tabular featurization requires apriori defining a finite number of classes per categorical parameter and maximum/minimum bounds for numbers for normalization -- when a new class of machines or workloads emerges, the entire process must be started from scratch and all the training data produced from the prior method's rigid featurization are made incomplete and invalidated.

\begin{figure}[htbp]
  \centering
  \begin{minipage}[t]{0.48\linewidth}
    \centering
    \vspace{-1.2cm}
    \tiny
    \begin{verbatim}
cell: cell_a
2024/06/02 17:00:00 PDT, Day:Fri Week:21
search_space:
  {'JOB/data_pipeline/PRODUCTION_WORKLOAD':
    ['machineE', 'machineA', 'none_selected']}
assignments: {"JOB/data_pipeline/PRODUCTION_WORKLOAD": "machineA"}
distributions:
- platform: {machineA}
  num_machines: 1.239e+03
  low_level_zones: 5.200e+01
  mid_level_zones: 5.200e+01
  high_level_zones: 4.300e+01
  resources: 5.481e+05
job_profiles:
- job: {user: data_pipeline, group_name: data_pipeline_workers}
  platform_profiles:
    machineD: {mean_mips_per_resource_usage: 8.165e+02}
    machineA: {mean_mips_per_resource_usage: 9.590e+02}
    machineF: {mean_mips_per_resource_usage: 8.321e+02}
    machineC: {mean_mips_per_resource_usage: 7.098e+02}
  limits:
    job_requested_resource_limit: 1.217e+04
    job_requested_num_vms: 1087
    \end{verbatim}
    \captionof{figure}{Example anonymized string representations of some features used to construct $x$. More detailed representation can be found in Appendix \ref{appendix:example_string}.}
    \label{fig:data_snippet}
  \end{minipage}
  \hfill
  \begin{minipage}[t]{0.48\linewidth}
    \centering
    \resizebox{\linewidth}{!}{%
      \begin{tabular}{lc}
        \hline
        Feature Type & Average Character Count \\
        \hline
        Cell Name & 3 \\
        Physical Location & 8 \\
        Time Window & 86 \\
        Scheduler Hyperparameters & 1,082 \\
        Machine Distribution & 461 \\
        Job-on-machine Performance & 268,157 \\
        \hline
      \end{tabular}%
    }
    \captionof{table}{Average character counts for each YAML feature.}
    \label{table:avg_char_counts}
    \vspace{2mm}
    \adjustbox{valign=t,clip,trim=0pt 0.2cm 0pt 0pt}{
    \includegraphics[width=\linewidth]{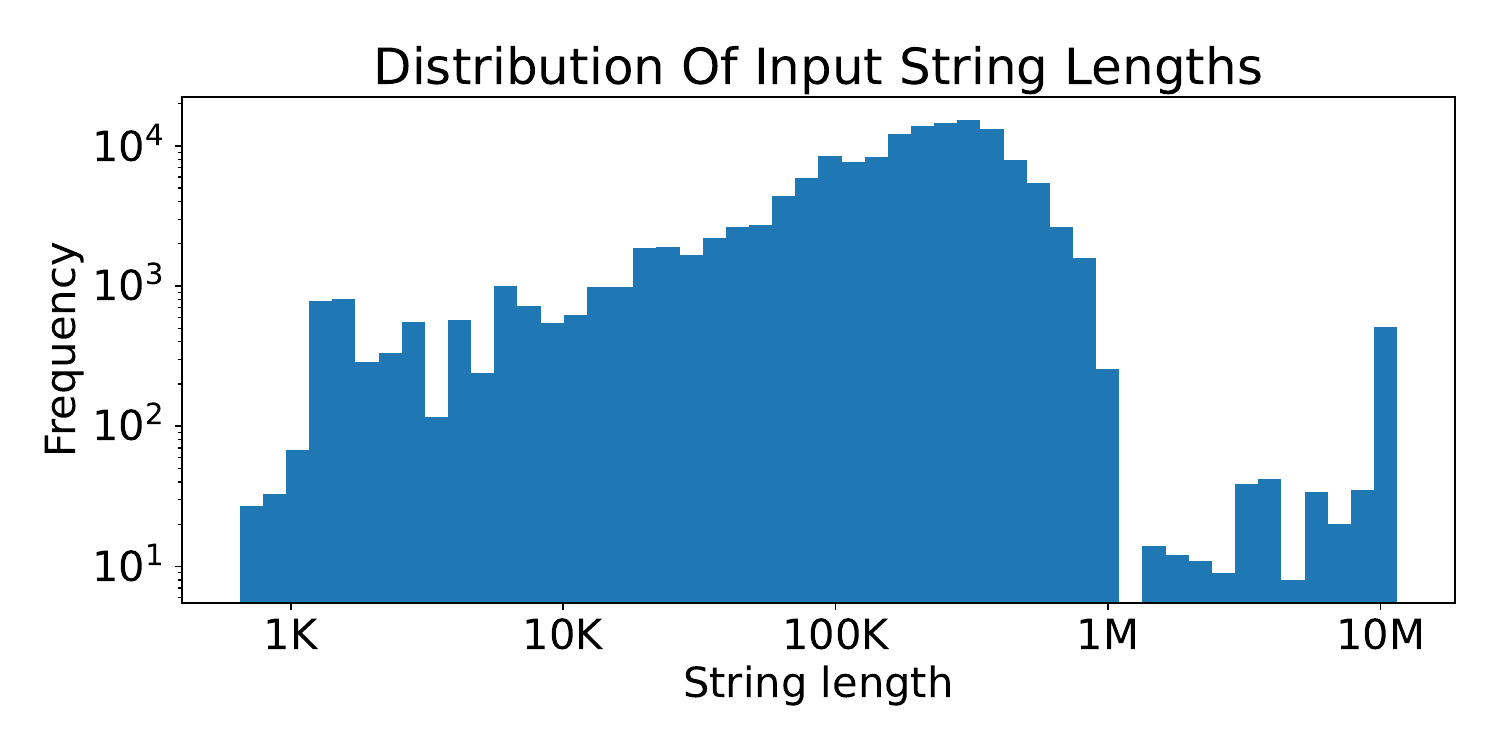}
    }
    \captionof{figure}{Distribution of string lengths.}
    \label{fig:stringlen_distribution}
  \end{minipage}
\end{figure}

A flexible text-based input representation simply resolves these issues by allowing variable sequence length inputs and does not require explicit enumeration of categorical features nor normalization of continuous ones. By observing all available features, especially nested ones difficult to represent as tabular formats, the model minimizes epistemic uncertainty and opens the doors to achieving the best performance possible. Furthermore, such a text-based model would not need to restart from scratch when encountering new sources of data, but rather can be simply fine-tuned over new tasks to allow fast few-shot adaptation by transferring knowledge learned from previously trained checkpoints.

In Figure \ref{fig:data_snippet}, we provide an example of a string, represented in a standard format (e.g. YAML), that can be sent to the model. Table \ref{table:avg_char_counts} gives the average character counts for each of the features across all data, while Figure \ref{fig:stringlen_distribution} gives the distribution of total string lengths.

\section{Method}
Below, we outline the technical details of our RLM, primarily based on OmniPred \citep{omnipred}. For a reference survey on standard language model training and fine-tuning, we refer the reader to \cite{zhang2023instruction}.

\subsection{Preliminaries: Text-to-Text Regression}
In the standard language model setting, given a batch of (prompt, response) pairs, one performs model updates by minimizing the next-token cross-entropy loss over response tokens. In the text-to-text regression case, the prompt corresponds to a string representation of $x$, while the response corresponds also to a textual or structured token representation of the floating value $y$. This training can be performed over \textit{any} collection of data points, due to the universality of string representations. At inference time, the model can be interpreted as a density estimator $p_{\theta}(y|x)$, from which a pointwise prediction $\widehat{y}$ can be made by aggregating i.i.d. samples $y^{(1)}, \ldots, y^{(s)} \sim p_{\theta}(y|x)$.

\subsection{Design Choices}
Some of the following design choices for our RLM can be considered contrary to popular belief. We justify their choices below, some of which require nuanced discussion based on findings from prior work and this paper's ablations:

\textbf{Decoding-based Output:} We train with cross-entropy loss over tokens rather than error-based loss (e.g. MSE) over an additional value head. The cross-entropy loss magnitude is agnostic to the strength of the prediction gap, leading to more stable training over multiple different $y$-values, and avoids over-focusing on tasks which naturally have larger errors due to wider $y$-value spreads. 

Furthermore, it has been broadly observed across both optimization \citep{universal_offline} and reward modeling communities \citep{generative_rm, generative_verifiers} that presumably due to their overly compressive nature, embedding-based or logit-based methods can perform worse than simply decoding the numeric prediction end-to-end.

\textbf{Use of Encoder:} Many current LLM designs only use the decoder-only architecture, as it simplifies designs by concatenating (prompt, response) pairs together without the need for specifying separate sequence lengths. However, as we show in our ablations in Section \ref{subsec:arch}, separate encoder layers are necessary for processing complex $x$, and decoder layers alone are suboptimal.

\textbf{No Language Pretraining:} Despite the current trend of pretraining models on \textit{human-generated text} in e.g. English, it is not necessary nor guaranteed beneficial to use a pretrained LLM checkpoint from which to train a regression model. In fact, \cite{omnipred} found that tabular regression is possible \textit{tabula rasa} with a randomly-initialized language model. This is presumably because regression only requires learning the correlations between different structured tokens and does not necessarily benefit from the semantic meaning behind words.

\textbf{y-Tokenization:} The cross-entropy loss does not have an inherent notion of numeric distance, as tokenization effectively discretizes the real number line. Since the model needs to learn an embedding for each token, it is important to minimize the vocabulary size used for representing floats, e.g. using <0>, <1>,..., <999> is far less effective than using a few digit tokens <0>, <1>, ..., <9>. We use the P10 tokenization found in \citep{p10}, in which a $y$ is represented using special sign, mantissa, and exponent tokens, e.g. <+><7><2><5><E-1> represents $725 \times 10^{-1} = 72.5$. This tokenization is also \textit{normalization-free}, which allows easy multi-task training without needing to precompute minimum or maximum $y$-value bounds for every separate task.

\textbf{Context-Free:} Our model is a \textit{context-free} regressor which only observes a single $x$ and returns a single $y$, instead of first preprocessing multiple $(x_1, y_1), (x_2, y_2), \ldots $ in-context \citep{llm_secretly_regressor}. This maximizes use of the sequence length for observing the string representation of a single $x$, which may already be thousands of tokens long. This further allows \textit{unlimited} data to be absorbed within the model weights, rather than having finite limits at inference time due to the context buffer. This distinction is analogous to using random forests and MLPs instead of Gaussian Processes for traditional regression, and weight-based updates instead of hidden memory states for meta-learning. 

\subsection{Fine-tuning}
Similar to standard LLM practices, the RLM also allows fine-tuning against additional data even after pretraining. This serves two different purposes: (1) adaptation to a new unseen regression task using pretrained knowledge, and (2) refocusing over a previously seen task (e.g. if the pretraining dataset was too large). 

To do so, we simply restore both the weights of a pretrained checkpoint and the optimizer state and resume standard training with a possibly lower learning rate, over the new data. The number of new examples can be arbitrarily low (e.g. 1-512), and effectively acts as a replacement to in-context learning via gradient update-based few-shot learning. Due to the relatively small size of our model, this procedure also only requires a few minutes on a single GPU.

This can be seen as a form of meta-learning \citep{maml} where pretraining leads to a checkpoint which can quickly be gradient-adapted to new tasks. Note that since the language model is able to observe a ``task-identifier feature'', e.g. the cell name, it can perform simultaneous regression over multiple tasks simply by pretraining over all task data without any specialized techniques, and then gain the ability to regress over new tasks after fine-tuning.

\subsection{Regression Scaling Paradigm}
The RLM maximizes scaling on multiple axes. However, we find that the two most important scaling factors for regression are diverse training data and feature observability. Since the task of regression is fundamentally based on learning the continuity of functions, e.g. that $y_1$ should be close to $y_2$ if $x_1$ and $x_2$ are also ``close'', better feature observability allows the model to learn better continuous representations of inputs $x$, while more training data provides better coverage over the input space.

In contrast, other axes such as model size are not necessarily very important; the task of regression is inherently discriminative and does not require large models for text generation. Furthermore, sequence length requirements can especially be reduced if a user with domain knowledge can efficiently compress string representations by e.g. removing commas or whitespaces, and also placing the presumably most important features at the beginning of the string representation. For this specific paper, our default model (exact details in Appendix \ref{appendix:experimental_settings}) only requires a 2 layer encoder-decoder (60M parameters) with 2K maximum sequence length for sufficient results.

\section{Experiments}
Overall, while our results are demonstrated on performance prediction for Google's Borg compute cluster, they are meant to serve an impression of what results may appear when text-to-text regression is applied to any large system. The general conclusions are:

\begin{itemize}
\item Maximizing feature observability enables significantly improved regression performance, surpassing previous baselines.
\item Large-scale pretraining on extensive datasets proves crucial, particularly for effective transfer learning and robust adaptation to new tasks.
\item RLMs prove to be ``universal'' in their capabilities, allowing both pointwise prediction and density estimation, while also scaling efficiently with data, model size, and sequence length.
\end{itemize}

\subsection{Data and Evaluation Protocol}
Specifically for our application, we define a \textit{task} $T$ as a collection of 28K-56K $\{(x_i, y_i)\}_{i \ge 0}$ state-outcome pairs specifically from a cell during a month (June or November). Pairs are randomly shuffled into train/validation/test splits in a standard 8/1/1 ratio. The model may be \textit{pretrained} on a combination of training splits from multiple tasks $\mathcal{T} = \{T_1, T_2, \ldots\}$, evaluated at test time either ``in-distribution'' for task $T'$ if $T' \in \mathcal{T}$, and ``out-of-distribution'' if $T' \notin \mathcal{T}$. Since each task is parameterized by cell and month, an out-of-distribution evaluation can be performed across either/both unseen time and cell axes, depending on the figure (exact details in Appendix \ref{appendix:experimental_settings}). To maximize benchmarking quality, our paper uses a total pool of 40 cells with the largest $y$-value spreads.


To evaluate regression performance, we use a variety of common measurements, appropriate to the analysis. MSE is used alongside lower bound $\text{TotalVariance}_{\phi}$ to assess precision. To avoid bias from different $y$-scalings between tasks, scale-invariant Spearman rank-correlation $(\rho)$ can also be used, especially when only rankings are sufficient, common in optimization-based applications. In ablation studies, we also use validation cross-entropy losses as simple yet accurate proxies of evaluation performance.


\input{experiments_main_v2}
\input{experiments_ablations}

\section{Conclusion}
We have validated the text-to-text regression approach for performance prediction in complex, industrial environments, by specifically demonstrating its effectiveness on Google's extensive compute cluster system. Our relatively cheap and simple encoder-decoder RLM, without relying on general language pretraining, can directly train over rich, non-tabular inputs like system logs and configuration files, return highly accurate floating point predictions, and quickly adapt to new tasks with minimal extra data.

Ultimately, this work showcases RLMs as powerful, general, and scalable tools for predicting metric outcomes from raw text. They alleviate the burdens of manual feature engineering and open new avenues for creating universal simulators for complex systems. Furthermore, by accurately modeling numeric feedback from varied inputs, RLMs can serve as a foundational aspect for developing sophisticated reward models to quickly give real-world feedback and operational ``experience'' \citep{era_of_experience}, catalyzing future research on reinforcement learning for language models.

\section*{Acknowledgements}
We would like to thank Uri Alon, Jonathan Lai, Mangpo Phothilimthana, Amir Yazdanbakhsh, David Smalling, Dara Bahri, Michal Lukasik, Rong-Xi Tan, Ke Xue, Shao-Hua Sun, Kuang-Huei Lee, Xinyun Chen, Chansoo Lee, Daiyi Peng, Jiyoun Ha, Aviral Kumar, Zi Wang, Gaurav Dhiman, and Yutian Chen for useful discussions and Yili Zheng, David Lo, Martin Dixon, Daniel Golovin, Denny Zhou, Claire Cui, Ed Chi, and Benoit Schillings for continuing support. 

\clearpage

\bibliography{references}
\bibliographystyle{abbrvnat}

\clearpage
\input{appendix}

\end{document}

%% file: math_commands.tex
\usepackage{amsmath,amsfonts,bm}

\def\eqref#1{equation~\ref{#1}}

\def\1{\bm{1}}

\DeclareMathAlphabet{\mathsfit}{\encodingdefault}{\sfdefault}{m}{sl}
\SetMathAlphabet{\mathsfit}{bold}{\encodingdefault}{\sfdefault}{bx}{n}



%% file: abbreviations.tex
\newcommand{\meA}{$C_{1}^{JUN}$}
\newcommand{\mgA}{$C_{2}^{JUN}$}
\newcommand{\mfA}{$C_{3}^{JUN}$}
\newcommand{\vzA}{$C_{4}^{JUN}$}
\newcommand{\lcphxpA}{$C_{5}^{JUN}$}
\newcommand{\roA}{$C_{6}^{JUN}$}
\newcommand{\rsA}{$C_{7}^{JUN}$}
\newcommand{\ggA}{$C_{8}^{JUN}$}
\newcommand{\nfA}{$C_{9}^{JUN}$}
\newcommand{\glA}{$C_{10}^{JUN}$}

\newcommand{\wmA}{$C_{22}^{JUN}$}
\newcommand{\tiA}{$C_{23}^{JUN}$}
\newcommand{\sgA}{$C_{24}^{JUN}$}
\newcommand{\ocA}{$C_{25}^{JUN}$}
\newcommand{\ddA}{$C_{26}^{JUN}$}
\newcommand{\itA}{$C_{27}^{JUN}$}
\newcommand{\voA}{$C_{28}^{JUN}$}
\newcommand{\slA}{$C_{29}^{JUN}$}
\newcommand{\waA}{$C_{30}^{JUN}$}
\newcommand{\lgA}{$C_{31}^{JUN}$}

\newcommand{\roC}{$C_{1}^{NOV}$}
\newcommand{\mfC}{$C_{2}^{NOV}$}

\newcommand{\mgC}{$C_{5}^{NOV}$}

\newcommand{\meC}{$C_{8}^{NOV}$}

\newcommand{\ukC}{$C_{10}^{NOV}$}
\newcommand{\vzC}{$C_{11}^{NOV}$}
\newcommand{\rsC}{$C_{12}^{NOV}$}
\newcommand{\ggC}{$C_{13}^{NOV}$}
\newcommand{\sgC}{$C_{14}^{NOV}$}

\newcommand{\itC}{$C_{16}^{NOV}$}
\newcommand{\dgC}{$C_{17}^{NOV}$}
\newcommand{\waC}{$C_{18}^{NOV}$}
\newcommand{\ilC}{$C_{19}^{NOV}$}

\newcommand{\ghC}{$C_{27}^{NOV}$}
\newcommand{\ddC}{$C_{28}^{NOV}$}

\newcommand{\voC}{$C_{30}^{NOV}$}
\newcommand{\tiC}{$C_{31}^{NOV}$}
\newcommand{\lgC}{$C_{32}^{NOV}$}
\newcommand{\lcphxpC}{$C_{33}^{NOV}$}
\newcommand{\slC}{$C_{34}^{NOV}$}

%% file: experiments_main_v2.tex
\subsection{Results}
\begin{figure}[t]
    \centering
    \includegraphics[width=\linewidth]{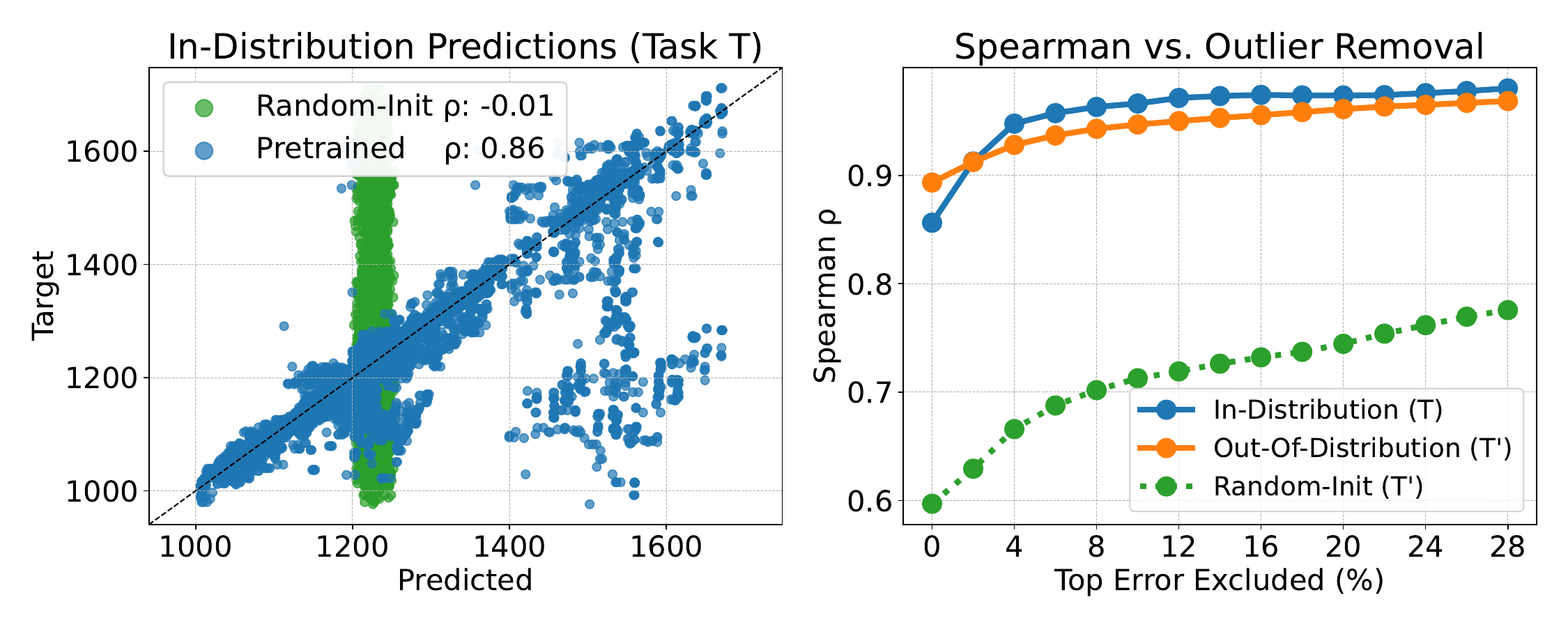}
    \caption{\textbf{Left:} Diagonal fit ($\diagup$) is better. RLM's pointwise prediction against ground truth target. \textbf{Right:} Higher is better ($\uparrow$). Spearman-rank correlation of the test evaluations, after removing top outliers by MSE error.}
    \label{fig:new_adaptation_mseexclusion}
\end{figure}

We begin with a case study on the RLM's ability to regress from strings given enough training data, on the task $T$ with the highest spread of $y$-values. In Figure \ref{fig:new_adaptation_mseexclusion} (left), we pretrain the RLM simultaneously over 8 tasks ($\approx$1M data points, or $\approx$2B tokens) and evaluate in-distribution on $T$ to achieve 0.86 rank correlation. Furthermore, when fine-tuning the pretrained checkpoint on only 512 few-shot examples from a completely new task $T'$, we see that the RLM can also achieve an equally strong result. In contrast, a randomly initialized checkpoint only given 512 examples is unable to achieve the same level of performance in either in-distribution and out-of-distribution cases, demonstrating the importance of large-scale pretraining and transfer learning.

\begin{figure}[h]
  \centering
    \centering
    \includegraphics[width=\textwidth]{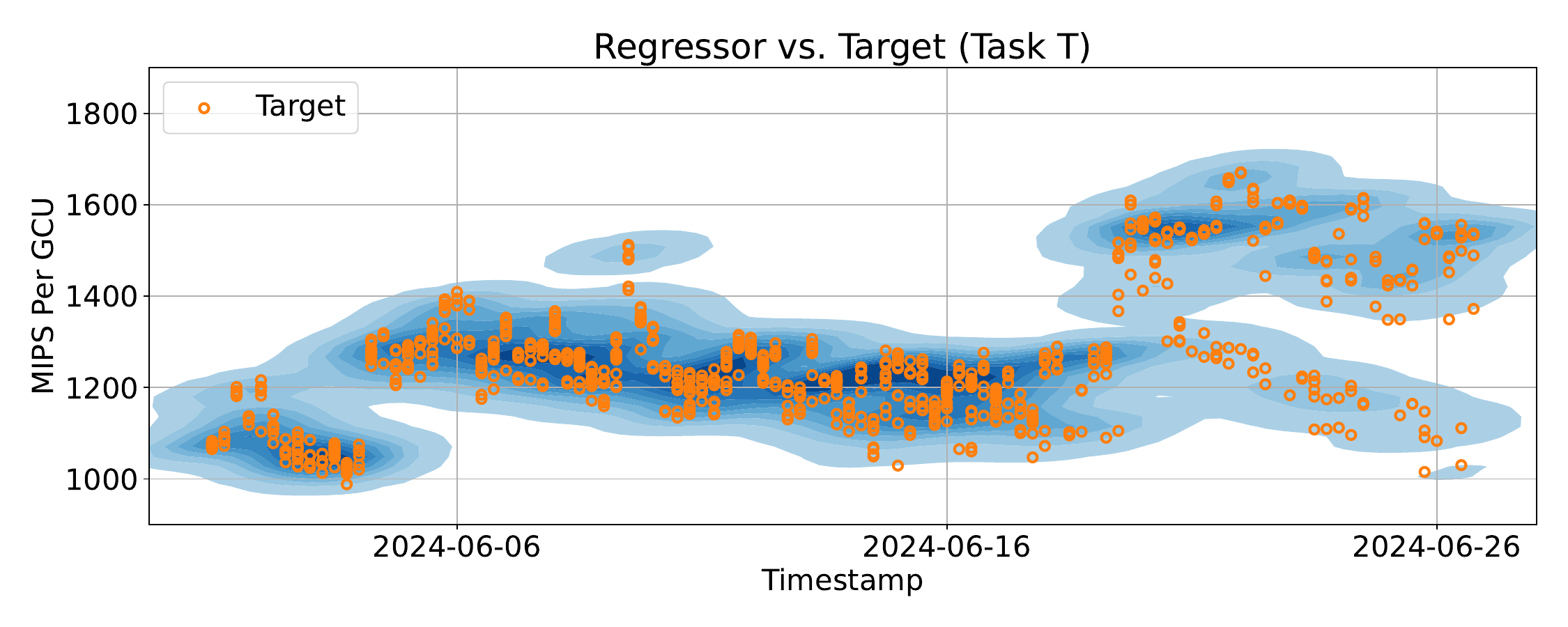}     
    \captionof{figure}{Better density capture of target points is better. Kernel Density Estimate (KDE) plot of samples from $p_{\theta}(y|x)$ along with actual target points, over varying timestamps. Note that samples are generated from $x$'s with distinct timestamps, while some target points may share timestamps.}
    \label{fig:new_idcell_kdecloud}
\end{figure}

Note that a large contribution to the outlier residual errors likely comes from aleatoric uncertainty, i.e. $y$ sampled from a probability distribution $p(y|x)$ rather than a deterministic pointwise outcome. Figure \ref{fig:new_idcell_kdecloud} also visualizes the regressor's density estimation abilities when varying along a single feature in $x$, i.e. the time when the computation was performed. Since the RLM's representation $\phi(x)$ is very rich and unique for every state $x$, the training data itself does not possess any multi-modality on $y$-values. However, it is remarkable that after training, the output $p_{\theta}(y|x)$ still expresses multiple modes, due to the RLM's inductive bias and learned continuous representations of $x$.

\begin{figure}[h]
    \centering
    \includegraphics[width=\linewidth]{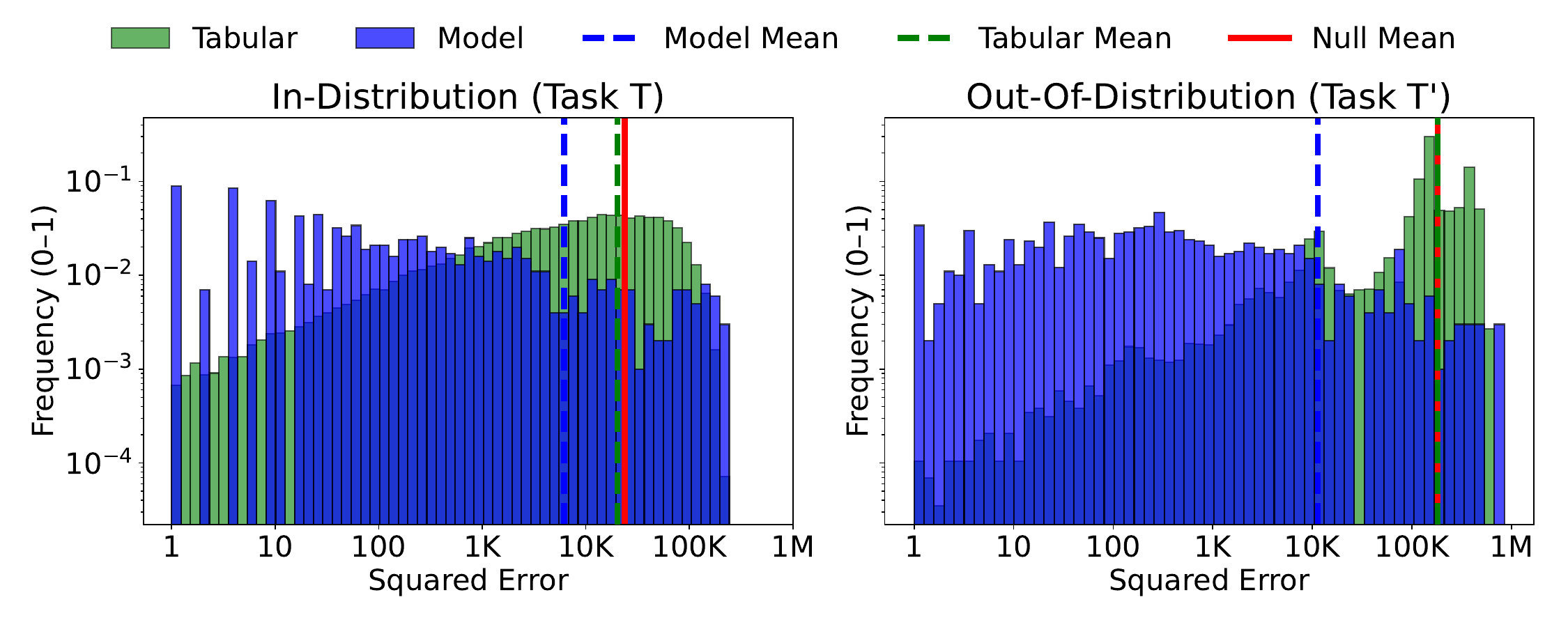}
    \caption{Left-skewness $(\leftarrow)$ is better. Note both axes are log-scaled. \textbf{Left:} Distribution of residuals as per-sample squared error, along with mean squared error as a vertical line. \textbf{Right:} Analogous results, but for an out-of-distribution task.}
    \label{fig:new_histograms_all}
\end{figure}

In Figure \ref{fig:new_histograms_all}, we further compare the RLM's performance against theoretically optimal results achievable by baselines using limited representations $\phi(x)$, i.e. observing only tabular hyperparameter features or nothing at all (``null''), using the $\text{TotalVariance}_{\phi}$ lower bounds in Section \ref{subsec:epistemic_uncertainty}. The RLM produces much lower residuals overall, with a 100x lower MSE than possible with tabular features, demonstrating the importance of maximizing feature observability.

\begin{figure}[h]
    \centering
    \includegraphics[width=\linewidth]{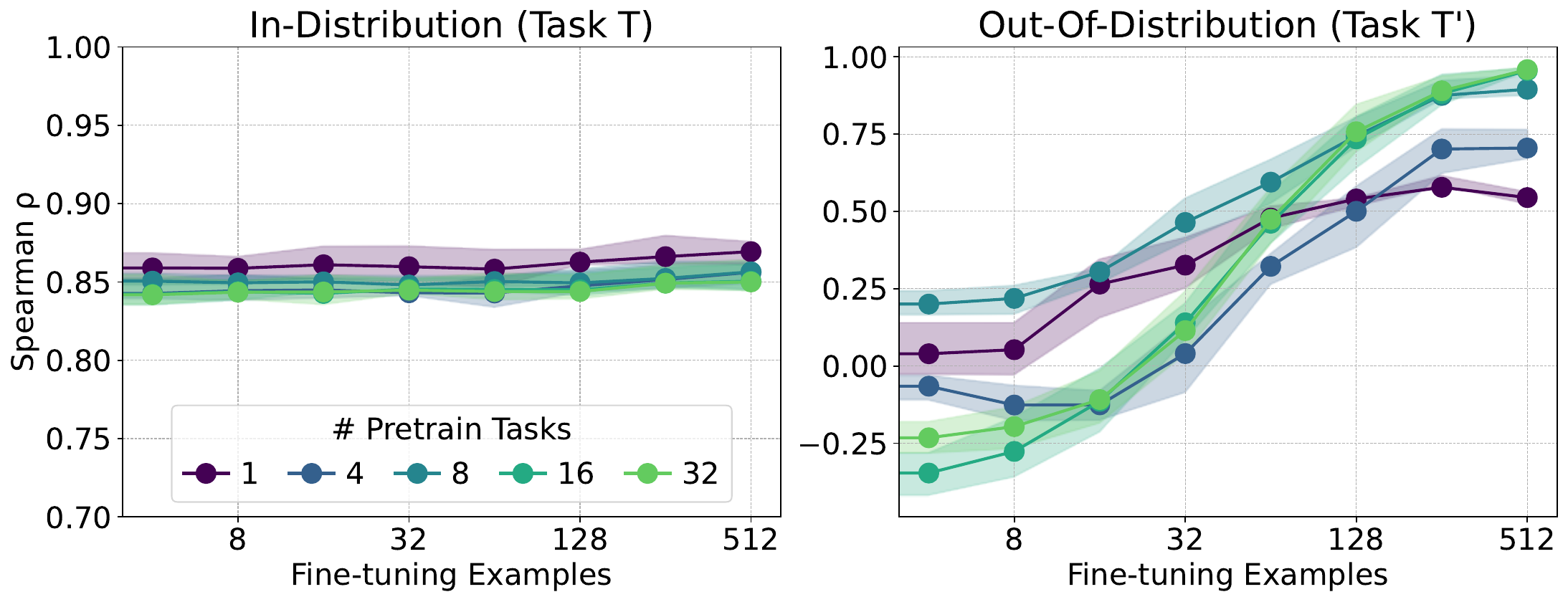}
    \caption{Higher is better ($\uparrow$). Evaluation task rank-correlation, when varying the number of finetuning examples starting from a pretrained checkpoint trained over varying numbers of tasks. Runs repeated over 10 seeds each and averaged.}
    \label{fig:main_finetuning}
\end{figure}

In Figure \ref{fig:main_finetuning}, we demonstrate the value of large-scale pretraining, especially for transfer-learning over new tasks. Our starter checkpoints are pretrained over a set of $N$ tasks for various $N$ over distinct cells. While the in-distribution case shows negligible gains when varying $N$, the out-of-distribution case shows that a higher $N$ leads to substantially better results on unseen cells.

\begin{figure}[ht!]
  \centering
  \begin{minipage}[t]{0.48\textwidth}
    \centering
    \includegraphics[width=\textwidth]{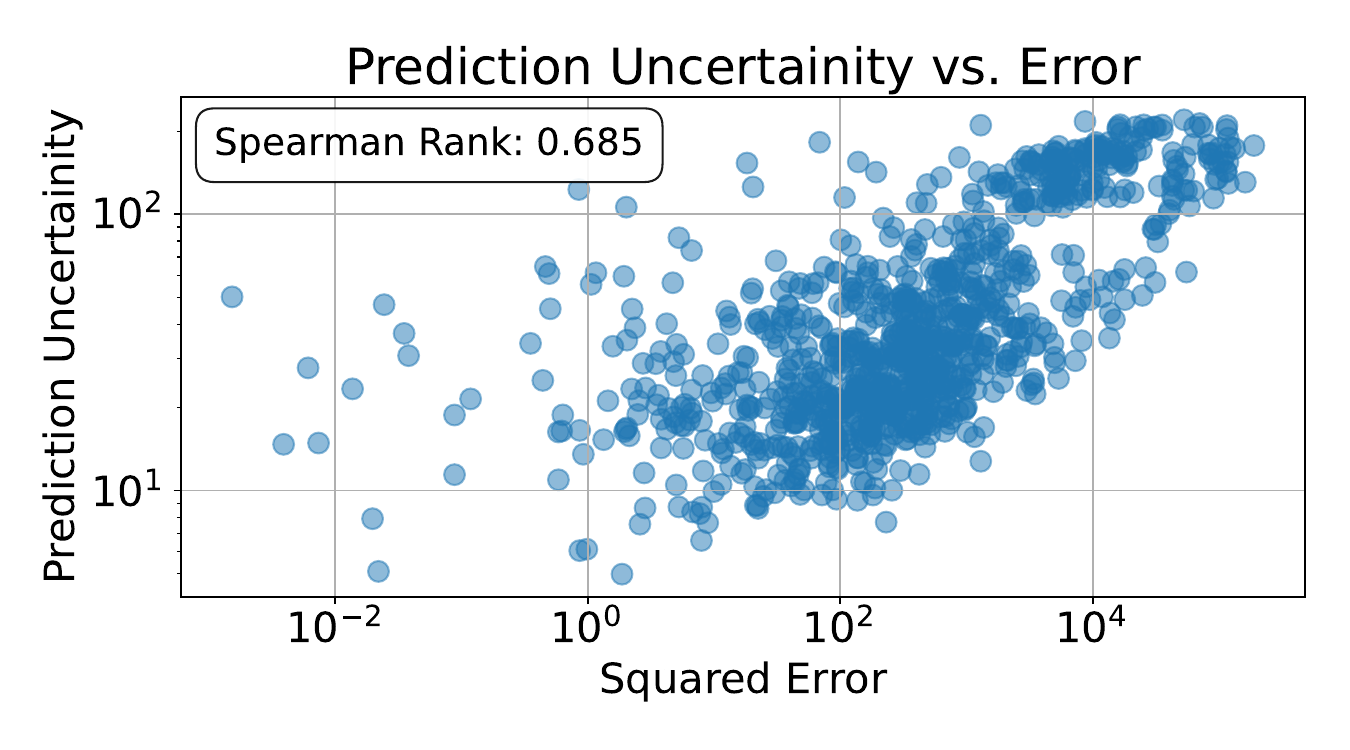}     
    \captionof{figure}{Correlation between prediction uncertainty (density variance) and residual squared error.}
    \label{fig:sample_uncertainity_msemean}
  \end{minipage}\hfill
  \begin{minipage}[t]{0.48\textwidth}
    \centering
    \includegraphics[width=\textwidth]{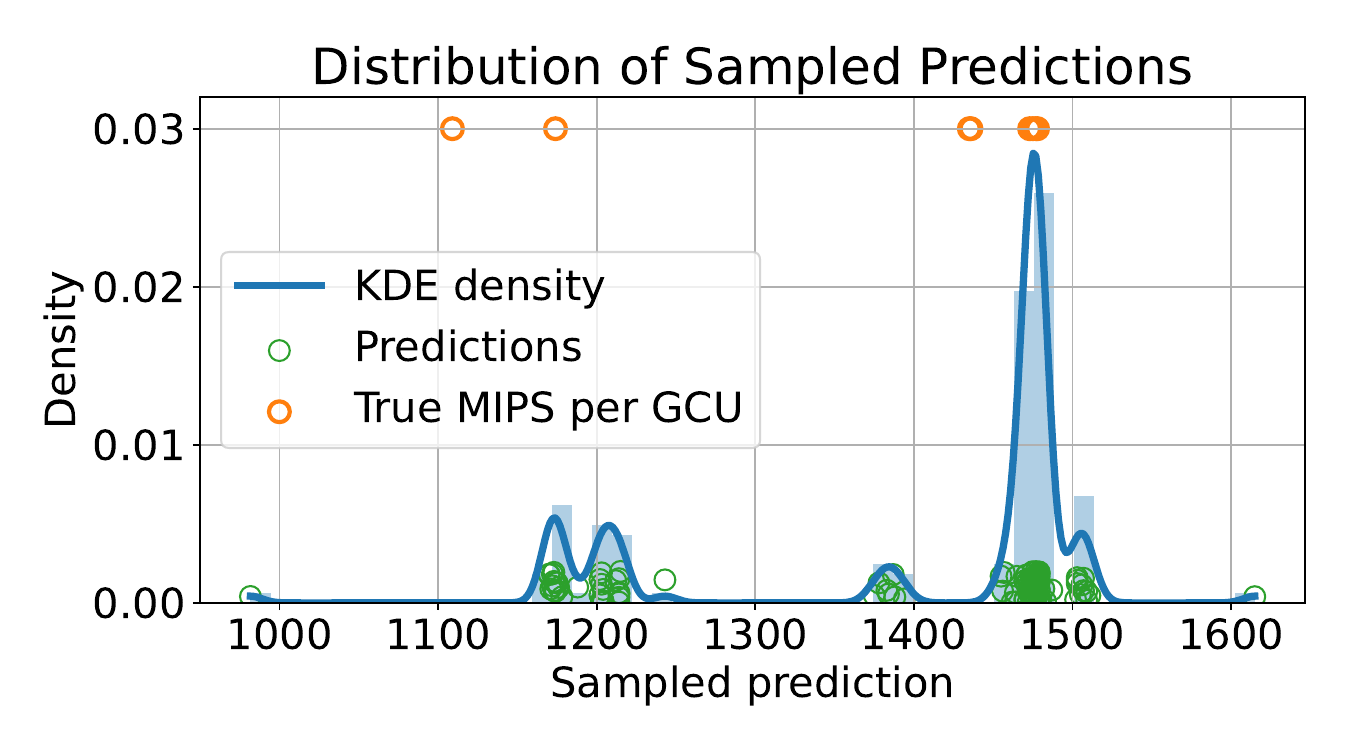}     
    \captionof{figure}{RLM density $p_{\theta}(y|x)$ for an example input $x$, along with true $y$-values.}
    \label{fig:bimodal_demo}
  \end{minipage}
\end{figure}

Regardless if the cause of error is from inherent randomness (aleatoric) or limited data coverage (epistemic), any regression method may fundamentally be inaccurate given certain inputs. For these cases, it should at least possess a higher uncertainty (e.g. density variance), crucial to downstream applications such as Bayesian Optimization. In Figure \ref{fig:sample_uncertainity_msemean}, we confirm there is a high correlation between the variance of $p_{\theta}(y|x)$ and the residual error from a sample $(x,y)$. Furthermore, in Figure \ref{fig:bimodal_demo}, we see that the RLM's density at a single example appropriately expresses bimodality when needed, corroborating Figure \ref{fig:new_idcell_kdecloud}.

While the above results show the RLM's useful properties, we finally present the overall comprehensive set of benchmarking results starting from a \textit{single} pretrained checkpoint, over various tasks denoted by $C_{cell}^{month}$ (see Appendix \ref{appendix:data_organization} for naming conventions). Multi-task results demonstrate the RLM's ability to perform inference simultaneously (when in-distribution) and via transfer learning (when out-of-distribution). Fundamentally, it is able to do so by identifying the task using the (cell, time) features and then observing other intra-task features to precisely predict the metric.

\begin{figure}[h]
    \centering
    \includegraphics[width=\linewidth]{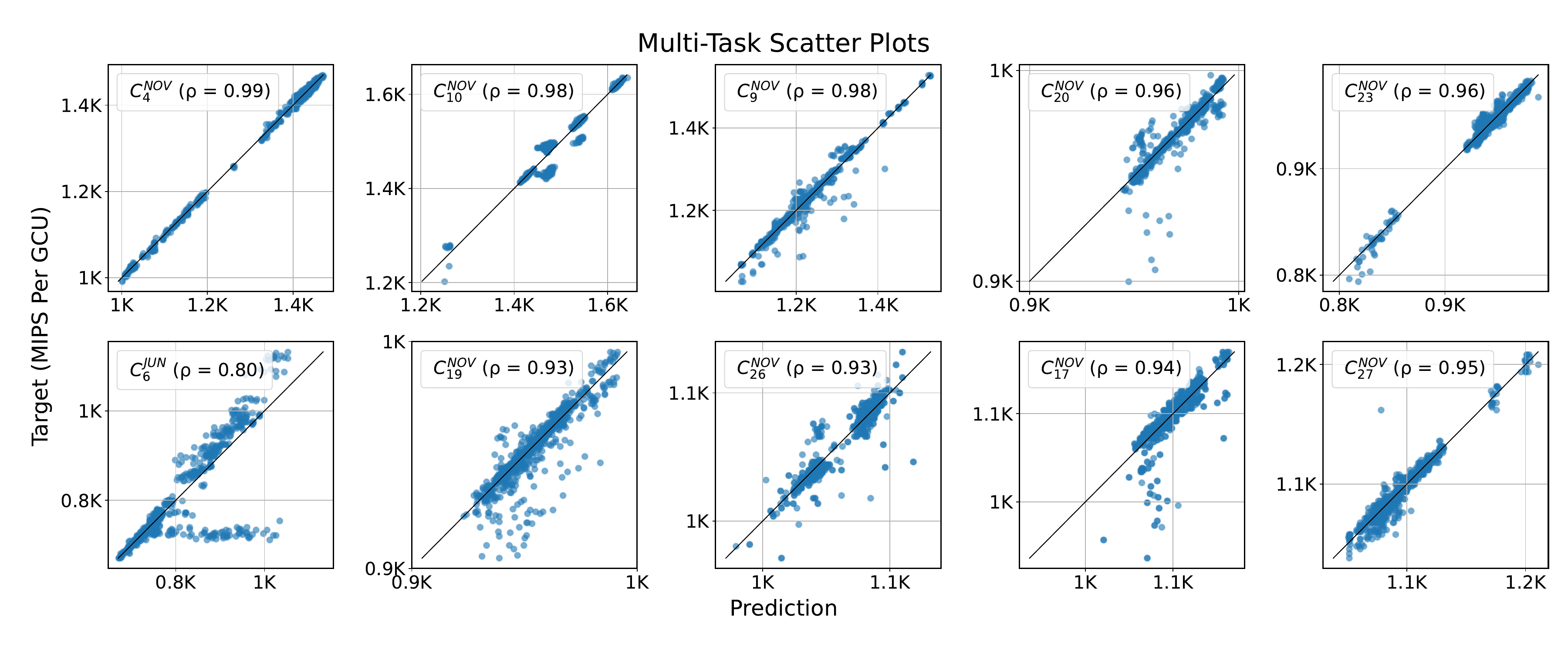}
    \caption{Diagonal fit ($\diagup$) is better. Scatter plot of predictions and ground truth targets over multiple tasks.}
    \label{fig:many_diagonals}
\end{figure}

We see that the RLM can achieve very precise pointwise regression (Figure \ref{fig:many_diagonals}) on a variety of tasks with different properties and scales, when the task presumably possesses low noise. For these cases, the Spearman rank is also very high, with the majority of cases reaching above 0.93.

\begin{figure}[h]
    \centering
    \includegraphics[width=\linewidth]{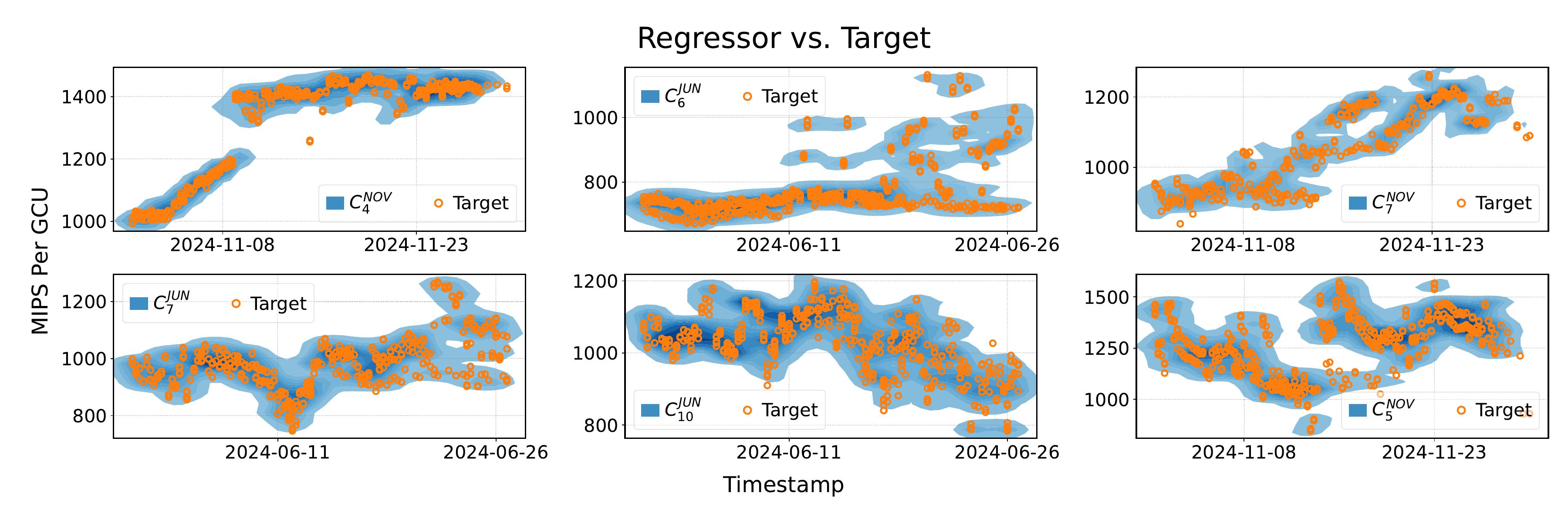}
    \caption{Better density capture of target points is better. KDE plot over multiple tasks.}
    \label{fig:many_kdeplots}
\end{figure}

Furthermore, in tasks likely to possess higher aleatoric noise which make pointwise predictions inappropriate, the RLM can perform density estimations instead (Figure \ref{fig:many_kdeplots}), capturing many different modes and density shapes.

\begin{figure}[h]
    \centering
    \includegraphics[width=\linewidth]{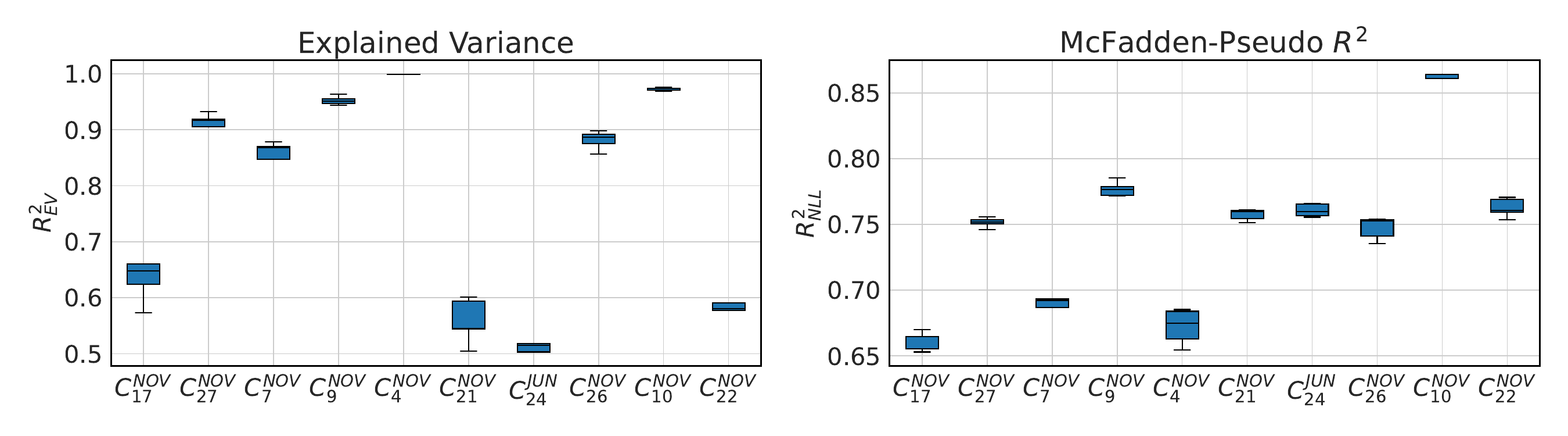}
    \caption{Higher is better $(\uparrow)$. \textbf{Left:} Explained variance per task. \textbf{Right:} McFadden's Pseudo-$R^2$ per task. Tasks are sorted by decreasing resultant rank correlations.}
    \label{fig:percell_finetuned_MSE_result}
\end{figure}

To quantify the actual gains of the RLM and to demonstrate the consistency of our technique, in Figure \ref{fig:percell_finetuned_MSE_result} we perform multiple runs and aggregate both the explained variance for pointwise regression and McFadden's Pseudo-$R^{2}$ \citep{mcfadden_rsquared} for density estimation, respectively denoted as $R^2_{\text{EV}} =  1 - \text{MSE}_{\text{RLM}} / \text{MSE}_{\text{null}}$ and $R^{2}_{\text{NLL}} = 1 - \text{NLL}_{\text{RLM}} / \text{NLL}_{\text{null}}$. Explained further in Appendix \ref{appendix:nll_r2}, these measurements capture the overall modeling gain of $p(y|x)$ from observing $x$ against the null case $p(y)$. Interestingly, the RLM achieves near perfect prediction on task $C_4^{NOV}$ with $R^{2}_{\text{EV}} \approx 1$, and even if the RLM achieves a lower $R^2_{\text{EV}}$ on noisy tasks, it can still achieve higher $R^{2}_{\text{NLL}}$ instead, as shown on tasks $C_{21}^{NOV}, C_{24}^{JUN}, C_{22}^{NOV}$.

%% file: experiments_ablations.tex
\section{Experiments: Ablations}

\subsection{Cross-Entropy Loss}
\begin{figure}[htbp]
  \centering
  \begin{minipage}[t]{0.48\textwidth}
    \centering
    \includegraphics[width=\textwidth]{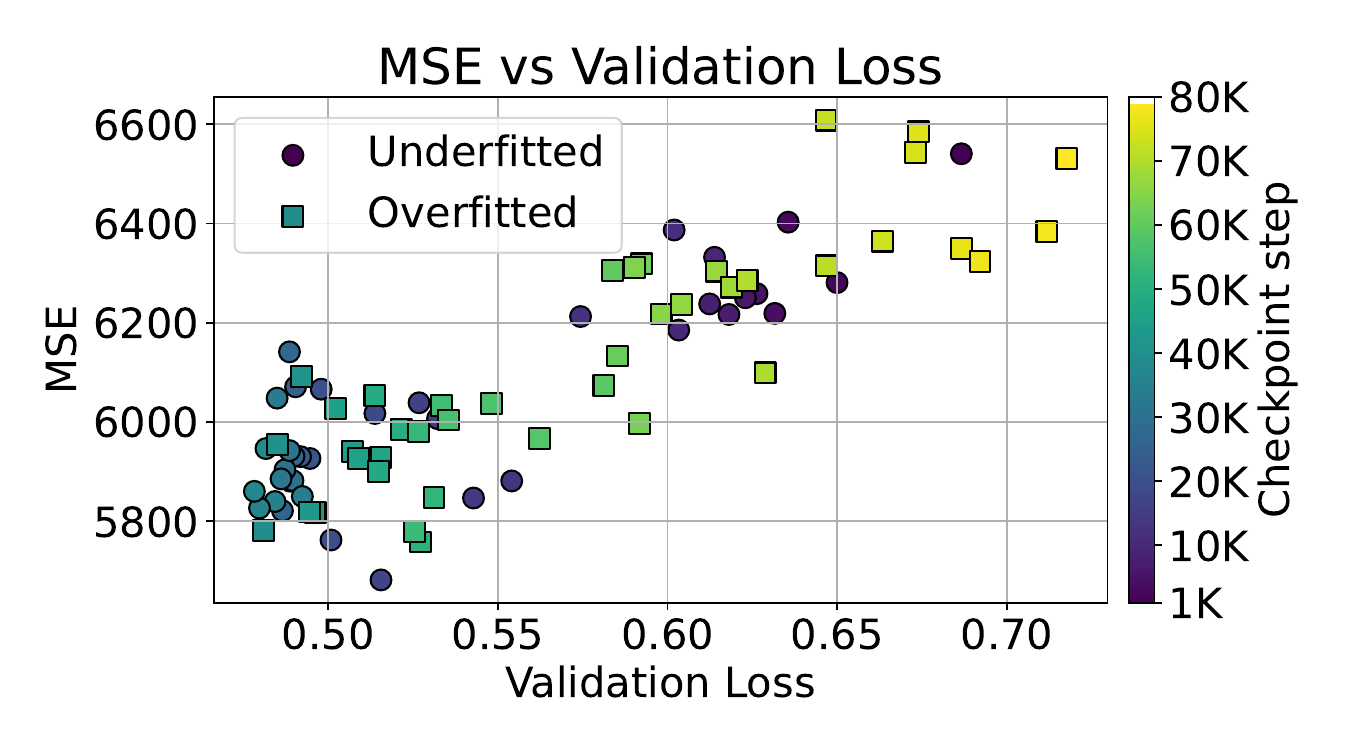}     
    \captionof{figure}{Lower is better $(\downarrow)$. MSE across checkpoint-steps.}
    \label{fig:ckpt_valloss_mse}
  \end{minipage}\hfill
  \begin{minipage}[t]{0.48\textwidth}
    \centering
    \includegraphics[width=\textwidth]{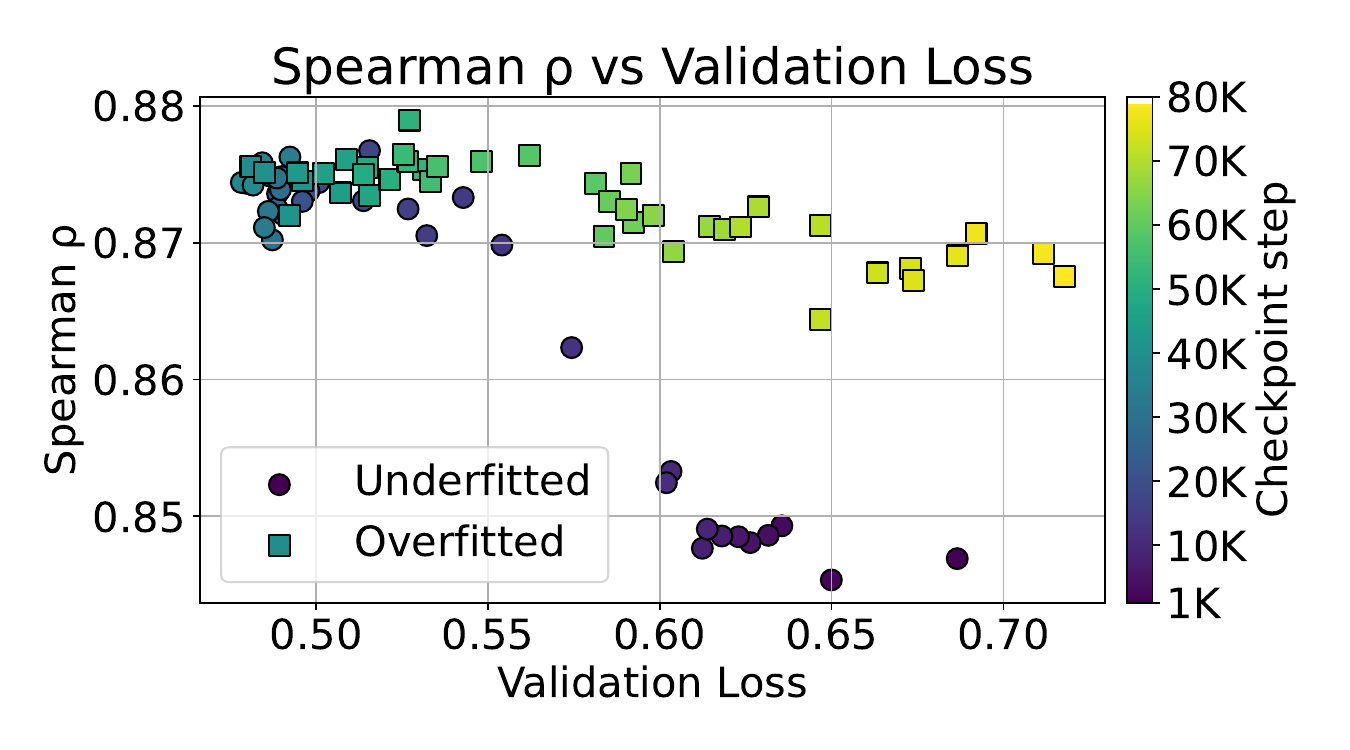}     
    \captionof{figure}{Higher is better $(\uparrow)$. Spearman $\rho$ across checkpoint-steps.}
    \label{fig:ckpt_valloss_spearmanrho}
  \end{minipage}
\end{figure}

We investigate the relationship between cross-entropy loss as a proxy for more traditional regression metrics such as MSE and Spearman rank correlation. During a training run, we evaluate all saved checkpoints, and based on the validation loss curve, label them as either underfitted or overfitted. In Figure \ref{fig:ckpt_valloss_mse}, we see a direct correlation with MSE, while interestingly in Figure \ref{fig:ckpt_valloss_spearmanrho}, we find that overfitted models can still maintain higher rank correlation.

\subsection{Architecture}
\label{subsec:arch}
\begin{figure}[htbp]
  \centering
  \begin{minipage}[t]{0.48\textwidth}
    \centering
    \includegraphics[width=\textwidth]{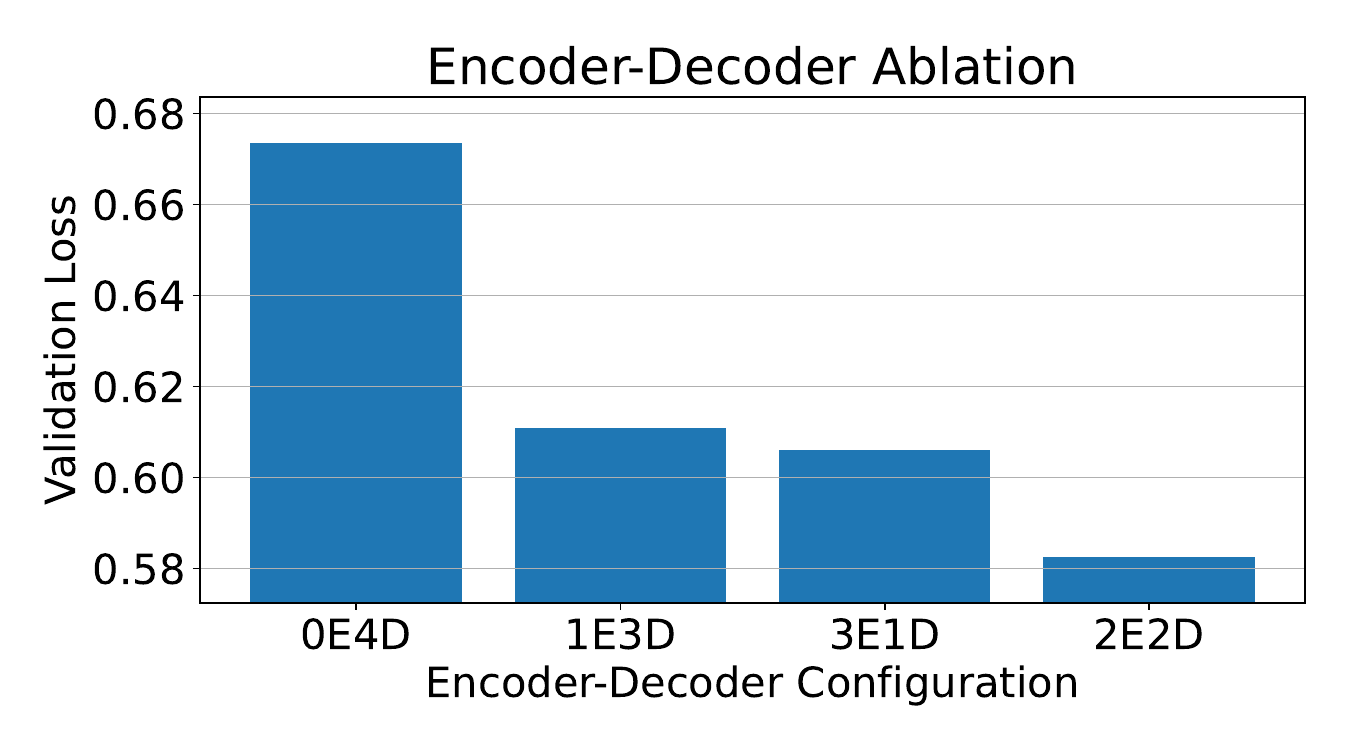}    
    \captionof{figure}{Lower is better $(\downarrow)$. Validation losses when varying architectures. Note: ``0E4D'' is a decoder-only model.}
    \label{fig:val_loss_encdec}
  \end{minipage}\hfill
  \begin{minipage}[t]{0.48\textwidth}
    \centering
    \includegraphics[width=\textwidth]{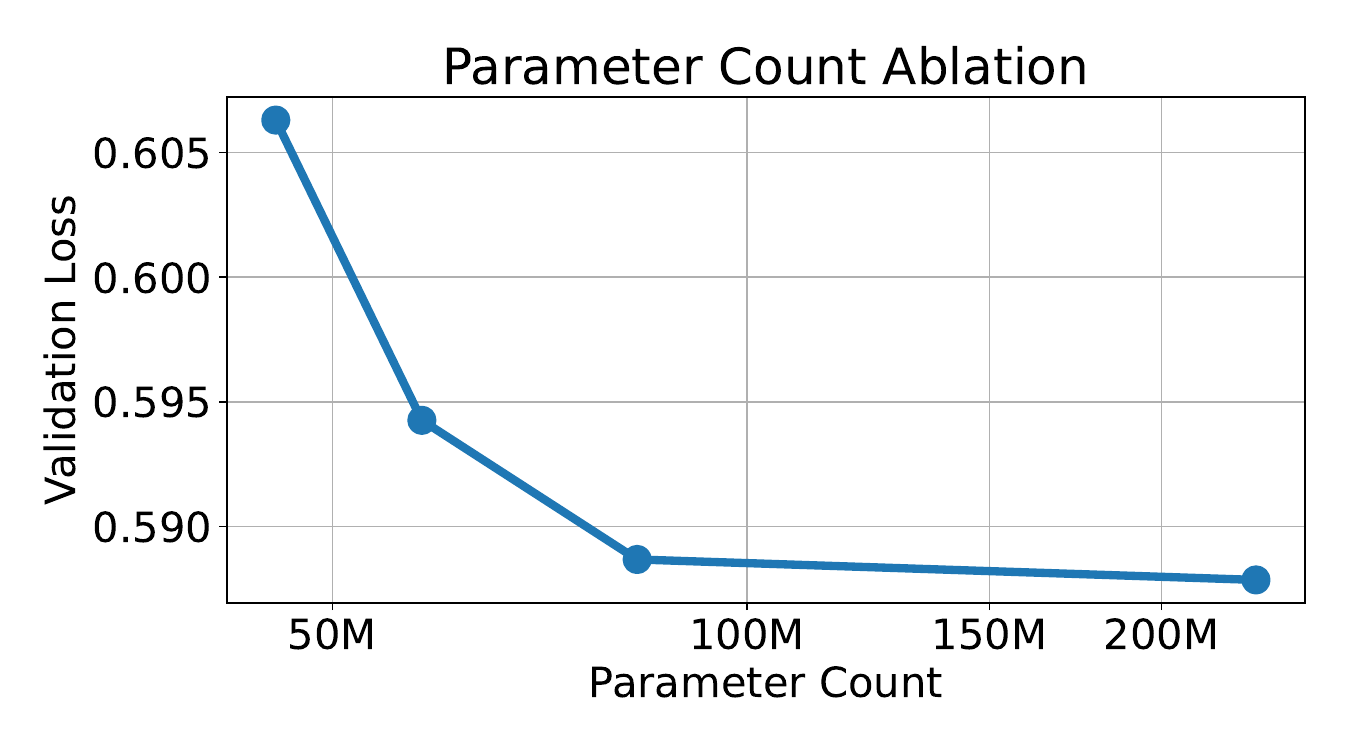}     
    \captionof{figure}{Lower is better $(\downarrow)$. Lowest observed validation losses when varying model sizes.}
    \label{fig:val_loss_param}
  \end{minipage}
\end{figure}

In Figure \ref{fig:val_loss_encdec}, we show the importance of processing the input $x$ with encoders. When accounting for equal model sizes (via layer count), we find that encoder-decoder architectures perform substantially better than decoder-only architectures even when using bidirectional attention on inputs. Note the stark contrast to LLM designs such as Gemma \citep{gemma} and Llama \citep{llama} which are decoder-only. We hypothesize that while decoder-only models are strong at producing outputs and chains of thought given relatively simple prompts, the information pathways routing through the decoder are insufficient to deal with complicated ``prompts'' $x$, although further investigation is needed.

In Figure \ref{fig:val_loss_param}, we also see that larger models with more parameter counts does improve regression performance, but surprisingly this quickly plateaus within the O(100M) range, which is orders of magnitudes lower than state-of-the-art general LLM models within the O(1B) range. This supports the broad utility of our text-to-text method, which requires relatively low amounts of compute, e.g. at most 1 GPU.

\subsection{Feature Importance}
\begin{figure}[htbp]
  \centering
  \begin{minipage}[t]{0.48\textwidth}
    \centering
    \includegraphics[width=\textwidth]{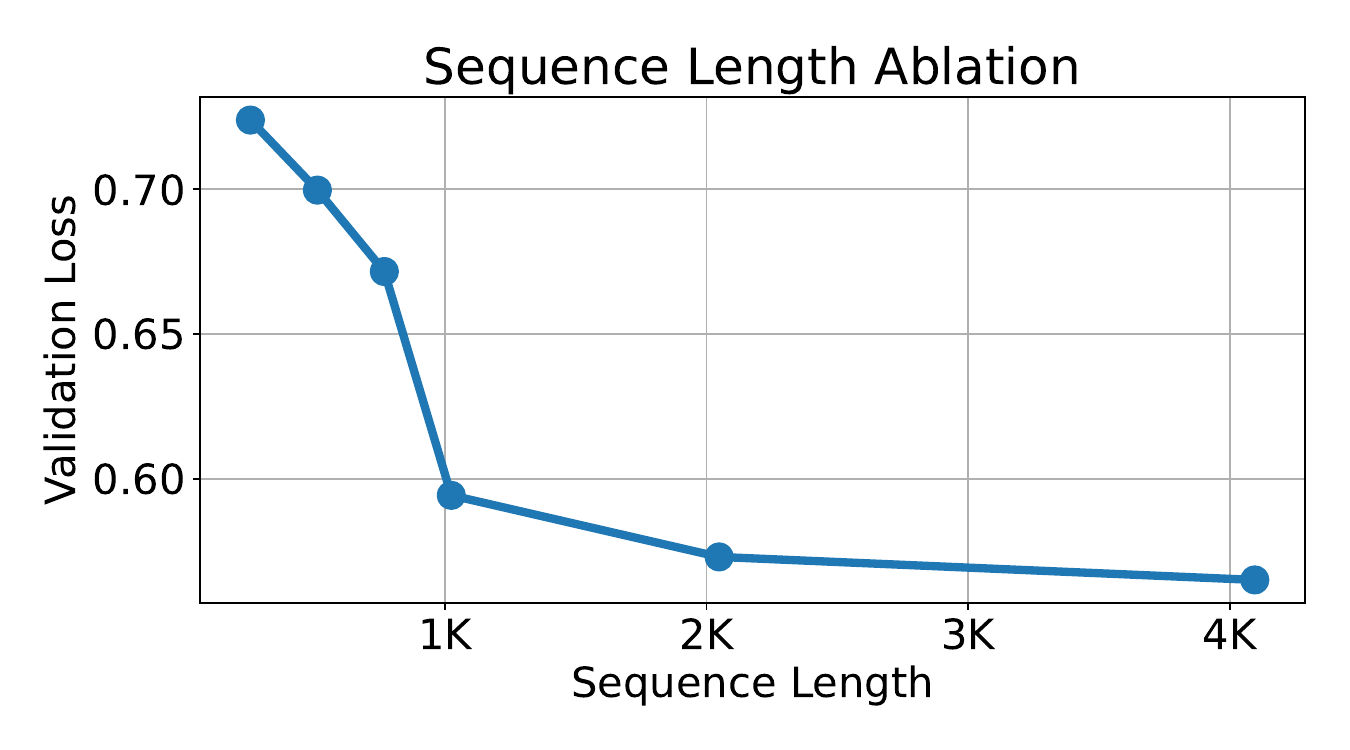} 
    \captionof{figure}{Lower is better $(\downarrow)$. Lowest observed validation losses when training over varying maximum sequence lengths.}
    \label{fig:val_loss_seq_len}
  \end{minipage}\hfill
  \begin{minipage}[t]{0.48\textwidth}
    \centering
    \includegraphics[width=\textwidth]{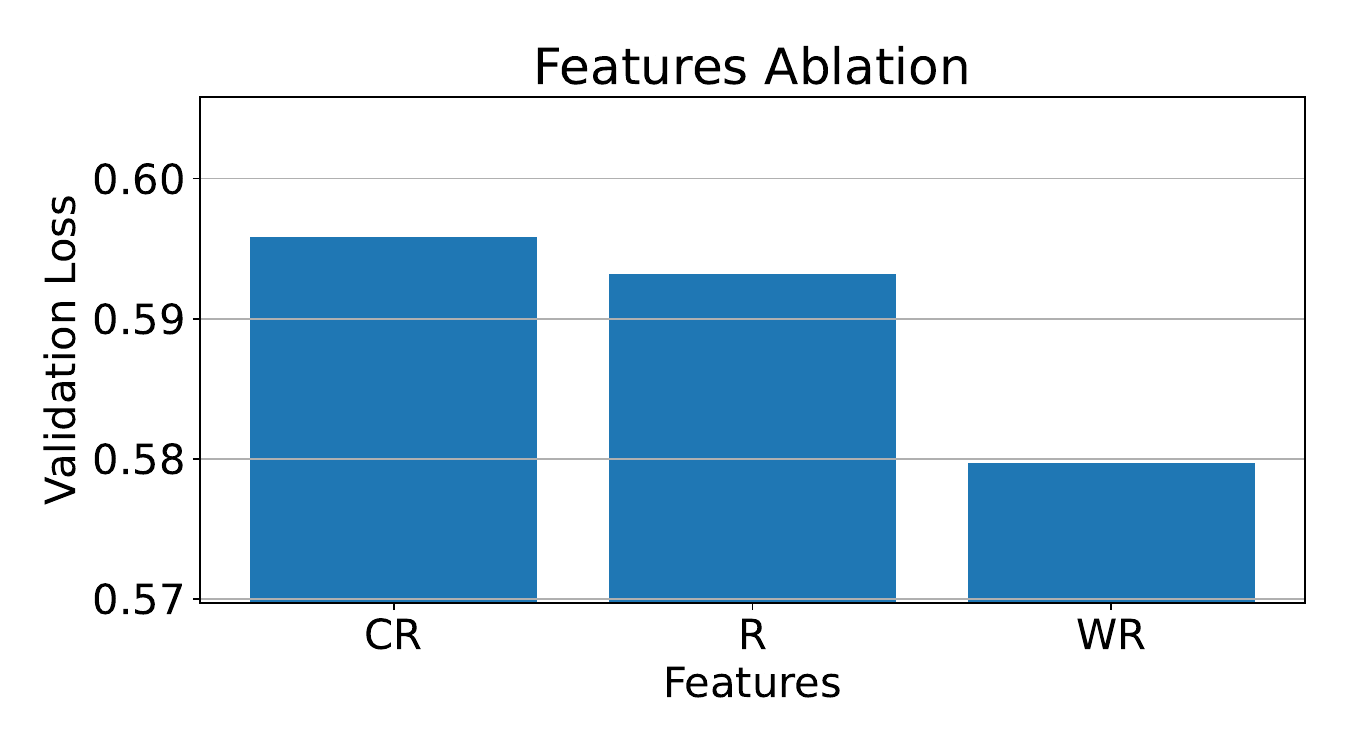}     
    \captionof{figure}{Lower is better $(\downarrow)$. Validation losses when showing certain features (``C'' = Cell, ``W'' = Window). ``R'' = using rest of the features.}
    \label{fig:val_loss_features}
  \end{minipage}
\end{figure}

In Figure \ref{fig:val_loss_seq_len}, we see that increasing the input length allows the model to observe more features from $x$ for predicting $y$, and thus achieving lower validation loss. When approximately $L \ge 3000$, the additional tokens mostly come from the last remaining, longest, and least important features, specifically the job-on-machine performance mentioned in Table \ref{table:avg_char_counts}, which explains the diminishing returns with higher sequence lengths.

In Figure \ref{fig:val_loss_features}, we also see that the model's behavior based on observing certain features aligns with our expectations from domain knowledge. For example, the performance of a cluster may heavily depend on temporal cycles -- there are fewer e.g. YouTube jobs at night due to lower user counts, or certain jobs may not run over the weekend. Experimentally, the model validates this intuition, as its performance substantially improves when its $x$ representation contains the period in which we performed bin-packing, i.e. time window feature, shown as e.g. \texttt{24-06-12T06:00:00Z} to \texttt{2024-06-12T06:05:00Z}.


\subsection{Few-Shot Adaptation}



\begin{figure}[htbp]
  \centering
  \begin{minipage}[t]{0.48\textwidth}
    \centering
    \includegraphics[width=\textwidth]{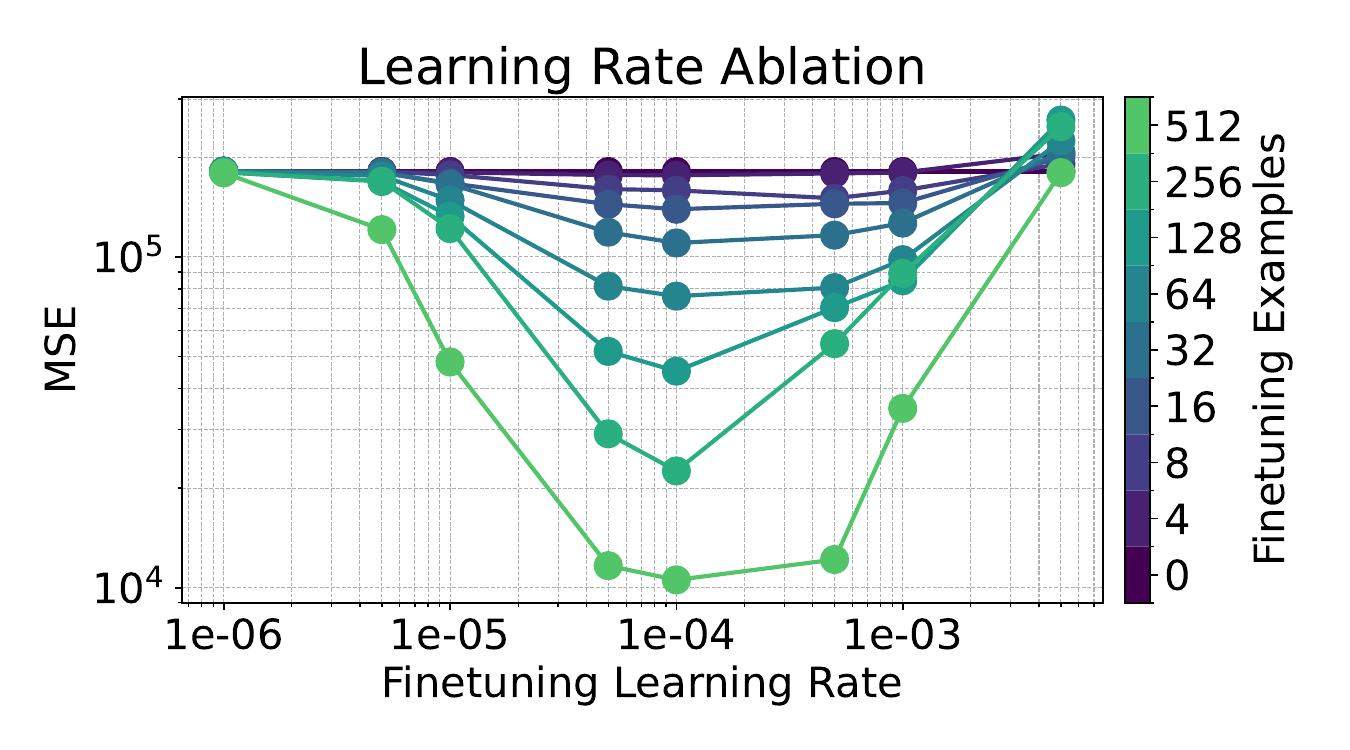}     
    \captionof{figure}{Lower is better $(\downarrow)$. MSE after OOD fine-tuning from a fixed checkpoint, while varying the learning rate. Early checkpoint (Step 10K) is used for adaptation.}
    \label{fig:mse_lr_ckpt_crossover}
  \end{minipage}\hfill
  \begin{minipage}[t]{0.48\textwidth}
    \centering
    \includegraphics[width=\textwidth]{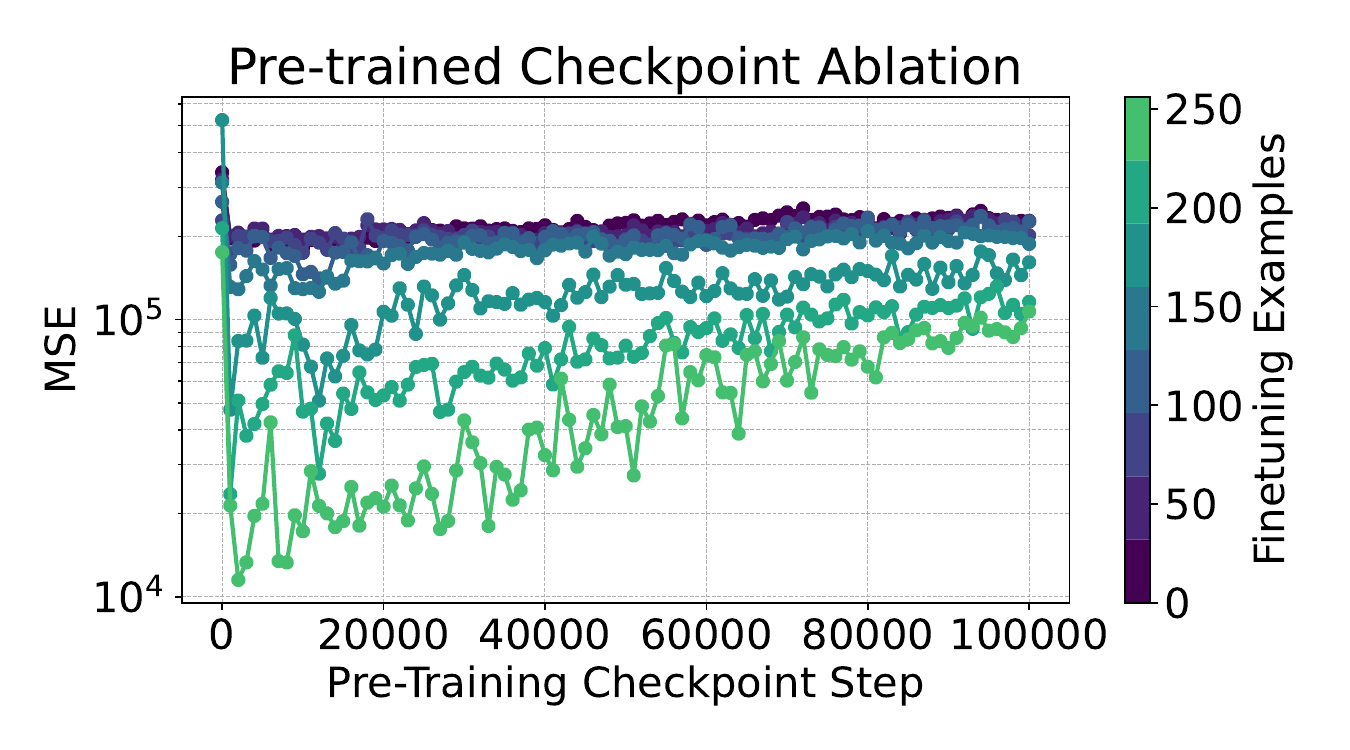}     
    \captionof{figure}{Lower is better $(\downarrow)$. MSE after OOD fine-tuning from a specific checkpoint-step during pretraining. A fixed learning rate of $5\times10^{-5}$ was used.}
    \label{fig:mse_checkpoint_scan}
  \end{minipage}
\end{figure}

Relevant for future practitioners, we provide useful ablations on the optimal settings for fine-tuning, and how they may affect results. In Figure \ref{fig:mse_lr_ckpt_crossover}, we see that finding the optimal learning rate matters significantly for reducing prediction errors when fine-tuning on multiple examples

In Figure \ref{fig:mse_checkpoint_scan}, we further see that earlier checkpoints from pretraining also can lead to better results when fine-tuning on out-of-distribution tasks. This is because an overly pretrained model may ``meta-overfit'' to the pretraining tasks themselves, making it harder to adapt to unseen tasks, especially those more different from pretraining.

%% file: appendix.tex
\appendix

\section{Additional Theory}

\subsection{Epistemic Uncertainty}
\label{appendix:epistemic}

Following up from Section \ref{subsec:epistemic_uncertainty} on partial observability, when only $\phi(x)$ is observed rather than the full state $x$, we can quantify the gap in terms of the well known Law of Total Variance:
\begin{equation}
\text{Var}(y|\phi(x)) = \mathbb{E}_{x \sim p(x|\phi(x))} \left[ \text{Var}(y|x) \right] + \text{Var}_{x \sim p(x|\phi(x))} \left( \mathbb{E}[y|x] \right)
\end{equation}
Here:
\begin{itemize}
\item $\mathbb{E}_{x \sim p(x|\phi(x))} \left[ \text{Var}(y|x) \right]$ is the average aleatoric uncertainty. It represents the expected value of the irreducible noise, averaged over all possible full states $x$ that are consistent with the observed partial representation $\phi(x)$.

\item $\text{Var}_{x \sim p(x|\phi(x))} \left( \mathbb{E}[y|x] \right)$ is the crucial term representing the contribution of epistemic uncertainty to the total variance. It is the variance of the true conditional mean $\mathbb{E}[y|x]$ (the optimal prediction if x were known) due to the unobservability of x given only $\phi(x)$. 
\end{itemize}

Since $\text{Var}_{x \sim p(x|\phi(x))} \left( \mathbb{E}[y|x] \right) \geq 0$, it follows that
\begin{equation}
\text{Var}(y|\phi(x)) \geq \mathbb{E}_{x \sim p(x|\phi(x))} \left[ \text{Var}(y|x) \right]
\end{equation}

This inequality formally shows that observing only a partial representation $\phi(x)$ increases (or at best, keeps the same) the variance that lower-bounds the regressor's achievable error, compared to the average aleatoric variance. The optimal performance of a regressor observing $\phi(x)$ is now limited by $\text{Var}(y|\phi(x))$. Thus, the limited observability captured by $\phi(x)$ induces epistemic uncertainty which directly contributes to a higher overall variance, further limiting the regressor's optimal performance.

\subsection{Explained Negative Log-Likelihood}
\label{appendix:nll_r2}
It is well known that the traditional explained variance $R^{2}_{\text{EV}}$ can be computed in terms of MSE ratios, i.e. $1 - \text{MSE}_{\text{RLM}} / \text{MSE}_{\text{null}}$, where the null case simply corresponds to the variance of the total $y$-population.

\cite{mcfadden_rsquared} also provides an analogous expression for density estimation cases, where negative log-likelihood (NLL) is instead used to calculate performance:
\begin{equation}
R_{\text{NLL}}^2  = 1 - \text{NLL}_{\text{RLM}} / \text{NLL}_{\text{null}}
\end{equation}
Note that simply using absolute $\text{NLL}_{\text{RLM}}$ as the final measurement is ambiguous, since there may not be an established reference for what is ``good'' or ``bad''. Instead, the reference is chosen as $\text{NLL}_{\text{null}}$ corresponds to a density estimator which models $y$ \textit{without} using $x$. Here, we choose the RLM itself, due to its decoder's universal density approximation abilities \citep{song2025decodingbasedregression} over more constrained distributions such as Gaussians. Note that $\text{NLL}_{\text{null}}$ is also an empirical approximation of the true entropy, which is impossible to calculate precisely as it requires online access to the true distribution.

\clearpage

\section{Additional Data Information}

\subsection{Example String Representation}
\label{appendix:example_string}

In Figure \ref{fig:data_snippet_full}, we show what the model exactly observes. Note the natural use of the newline character ``\verb|\n|'' as a natural separator between different features.

\begin{figure}[h]
    \centering
    \lstset{
        basicstyle=\ttfamily\tiny,
        breaklines=true,
        breakatwhitespace=false,
        breakindent=0pt,
        columns=fullflexible, 
    }
    \begin{lstlisting}
'\ncell: cell_a\n\n\'2024-06-12T06:00:00Z\'\n\n\'2024-06-12T06:05:00Z\'\n\nLarge users: [\'datapipeline-prod\', \'researcher_human_1\', \'monitoring-jobs\', \'model-server-public-prod-jobs\', \'ecommerce\', \'ads_fetch\', \'researcher_human_2\', \'real-time-diagnostics\', \'storage_1\', \'storage_2\', \'storage_3\', \'webserver\', \'storage_aggregation\', \'storage_frontend\', \'researcher_human_5\', \'routing\', \'storage_4\', \'researcher_human_4\', \'storage_5\', \'analytics-storage_6\']\nsearch_space: {\'machineA\': [\'machineA\', \'none_selected\']} {\'machineD,machineC\': [\'machineD\', \'machineC\', \'none_selected\']} {\'machineA,machineE\': [\'machineA\', \'machineE\', \'none_selected\']} {\'machineF\': [\'machineF\', \'none_selected\']} {\'machineD,machineA\': [\'machineD\', \'machineA\', \'none_selected\']} {\'machineD\': [\'machineD\', \'none_selected\']} {\'machineE\': [\'machineE\', \'none_selected\']} {\'machineD,machineE\': [\'machineD\', \'machineE\', \'none_selected\']} {\'JOB/storage_4/PRODUCTION_WORKLOAD\': [\'machineE\', \'none_selected\']} {\'JOB/processing_1/me.processing_1.processing_1/PRODUCTION_WORKLOAD\': [\'machineE\', \'none_selected\']}\n\nassignments: {"JOB/storage_4/PRODUCTION_WORKLOAD": "none_selected", "JOB/processing_1/me.processing_1.processing_1/PRODUCTION_WORKLOAD": "machineE", "machineD": "machineD", "machineD,machineA": "machineA", "machineD,machineE": "machineD", "machineD,machineC": "none_selected", "machineA": "machineA", "machineA,machineE": "machineE", "machineF": "machineF", "machineE": "none_selected"}\ncell: cell_a\ndistributions:\n- platform: {machineA}\n  num_machines: 1.239e+03\n  low_level_zones: 5.200e+01\n  mid_level_zones: 5.200e+01\n  high_level_zones: 4.300e+01\n  resources: 5.481e+05\n- platform: {machineE}\n  num_machines: 1.070e+02\n  low_level_zones: 3.900e+01\n  mid_level_zones: 3.900e+01\n  high_level_zones: 3.500e+01\n  resources: 2.430e+04\n- platform: {machineD}\n  num_machines: 5.640e+02\n  low_level_zones: 3.000e+01\n  mid_level_zones: 3.000e+01\n  high_level_zones: 1.900e+01\n  resources: 2.769e+05\n- platform: {machineC}\n  num_machines: 6.100e+01\n  low_level_zones: 6.000e+00\n  mid_level_zones: 6.000e+00\n  high_level_zones: 3.000e+00\n  resources: 2.306e+04\n- platform: {machineF}\n  num_machines: 1.880e+02\n  low_level_zones: 1.200e+01\n  mid_level_zones: 1.200e+01\n  high_level_zones: 6.000e+00\n  resources: 4.527e+04\n\njob_profiles:\n- job: {user: leaf, group_name: bun}\n  platform_profiles:\n    machineD: {mean_productivity_per_resource_usage: 1.048e+00, mean_mips_per_resource_usage: 8.588e+02}\n    machineF: {mean_productivity_per_resource_usage: 1.060e+00, mean_mips_per_resource_usage: 8.285e+02}\n    machineE: {mean_productivity_per_resource_usage: 1.100e+00, mean_mips_per_resource_usage: 8.292e+02}\n    machineC: {mean_productivity_per_resource_usage: 9.721e-01, mean_mips_per_resource_usage: 7.911e+02}\n  observed_resource_limit: 7.861e+02\n  observed_num_vms: 9.000e+00\n  limits: {machine_limit: 1.000e+00, switch_limit: 1.000e+00, low_level_limit: 1.000e+00, has_port_limit: true, job_requested_resource_limit: 8.832e+02, job_requested_num_vms: 10, group_requested_resource_limit: 8.832e+02}\n  productivity_per_resource_1: 1.048e+00\n  productivity_per_resource_2: 1.045e+00\n- job: {user: cent, group_name: rock}\n  platform_profiles:\n    machineD: {mean_productivity_per_resource_usage: 3.803e+01, mean_mips_per_resource_usage: 5.716e+02}\n    machineA: {mean_productivity_per_resource_usage: 7.685e+00, mean_mips_per_resource_usage: 4.902e+02}\n    machineE: {mean_productivity_per_resource_usage: 5.893e+01, mean_mips_per_resource_usage: 5.610e+02}\n    machineC: {mean_productivity_per_resource_usage: 3.274e+01, mean_mips_per_resource_usage: 7.173e+02}\n  observed_resource_limit: 6.400e+02\n  observed_num_vms: 2.000e+01\n  limits: {job_requested_resource_limit: 6.400e+02, job_requested_num_vms: 20, group_requested_resource_limit: 6.400e+02}\n  productivity_per_resource_1: 1.941e+01\n  productivity_per_resource_2: 3.435e+01\n- job: {user: pond, group_name: green}\n  platform_profiles:\n    machineD: {mean_productivity_per_resource_usage: 1.413e+06, mean_mips_per_resource_usage: 7.420e+02}\n  observed_resource_limit: 1.100e+02\n  observed_num_vms: 1.900e+01\n  limits: {job_requested_resource_limit: 1.362e+04, job_requested_num_vms: 1401, group_requested_resource_limit: 1.362e+04}\n  productivity_per_resource_1: 1.413e+06\n  productivity_per_resource_2: 1.413e+06\n
    \end{lstlisting}
    \caption{Example anonymized string representation of $x$, truncated to fit.}
    \label{fig:data_snippet_full}
\end{figure}

\subsection{Data Organization and Statistics}
\label{appendix:data_organization}
\begin{figure}[h]
    \centering
    \includegraphics[width=\linewidth]{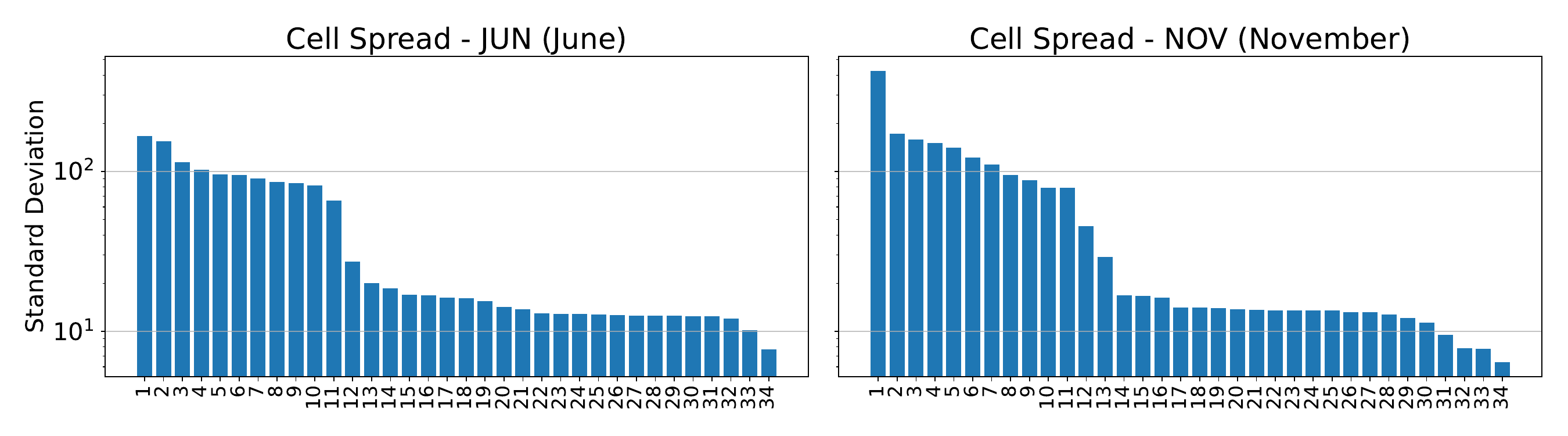}
    \caption{Spread of cells included in our study, separated by month.}
    \label{fig:variance_percell}
\end{figure}

Due to corporate privacy concerns, we are required to anonymize all cell names, but make sure the naming scheme is consistent across all figures. Each task is labeled as $C_{cell}^{month}$, parameterized by cell (integer index) and month (name). The index of a cell is sorted in reverse order according to its ``spread'' of $y$-values, i.e. variance of $p(y)$ unconditioned on $x$, and we show the distribution of spreads in Figure \ref{fig:variance_percell}. Thus \meA and \mgA within June and \roC and \mfC within November are the highest-spread cells, with June cells having dataset sizes 54K-56K and November having 28K.

\clearpage

\input{experimental_description_appx}

%% file: experimental_description_appx.tex
\section{Experimental Settings}
\label{appendix:experimental_settings}

\subsection{Default Settings}
We used the standard T5X EncoderDecoder which can found in the open-source
codebase \url{https://github.com/google-research/t5x.} Below are default settings (unless otherwise overridden). 

For training/pretraining:



\begin{itemize}
    \item \underline{Optimizer:} Adafactor \citep{adafactor} with base learning rate $0.1$ and square root decay with $1000$ warm-up steps, $0.5$ decay factor, $2000$ steps per decay, and $10000$ steps per cycle. Batch size 128 with 8 microbatches.
    \item \underline{Vocabulary and Tokenizer:} SentencePiece tokenizer \citep{sentencepiece} with T5X's default vocabulary of 32000 subword tokens, in addition to the custom P10 tokens \citep{p10} for representing $y$-objectives, with 4-digit mantissas.
    \item \underline{Early stopping:} We train for a maximum of 100K steps, but early stop based on validation loss if overfitting is detected.
    \item \underline{Architecture:} 2 encoder layers, 2 decoder layers, 16 heads, 64 head dimension, 512 embedding dimension, 2048 MLP dimension, leading to 58M parameters. Default sequence length 2048, with truncation occurring afterwards.
\end{itemize}

For fine-tuning:

\begin{itemize}
    \item \underline{Optimizer:} Restore checkpoint optimizer state, but change the learning rate to $5 \times 10^{-5}$.
    \item \underline{Batch size:} 128. Note that if the finetuning data size is lower, we simply repeat up to this size.
    \item \underline{Early stopping:} Perform up to 200 epochs of finetuning, but early stop based on validation loss if overfitting is detected.
\end{itemize}

For inference:

\begin{itemize}
    \item \underline{Sampling:} 128 parallel decoding examples per single $x$, while removing accidental "extreme" outliers outside of the known range [500, 3000] of possible achievable MIPs per GCU values.
    \item \underline{Pointwise Aggregation:} Take the numeric mean of samples (to minimize MSE) or numeric median (to minimize Spearman rank).
\end{itemize}

\subsection{Main Training Runs}

All of the figures in this paper stem from a few main training setups, focused on evaluating on both in-distribution and out-of-distribution cases, or ablations. Specifically, these main settings:

\begin{itemize}
\item \underline{Limit Testing:} We train a larger model (267M params) with 4096 sequence length and batch size 256, over nearly all tasks except the highest spread cells across the two months $\{C_1^{JUN}, C_1^{NOV}\}$. We then also evaluate on the highest spread cells possible across the two months, i.e. $C_{2}^{JUN}$ (in-distribution) and $C_{1}^{NOV}$.
\item \underline{Adaptation Testing:} We train five separate checkpoints over $\{1,4,8,16,32\}$ different tasks chosen randomly, but making sure there is a uniquely seen task $C_2^{JUN}$ by all checkpoints, and also evaluate on a out-of-distribution task $C_{10}^{NOV}$ unseen by all checkpoints.
\item \underline{Ablations:} We simply pretrain on 7 high spread tasks from June.
\end{itemize}

More specifically, the pretraining tasks for \underline{Limit Testing} are the combination of \mgA to \glA, \wmA to \lgA, \mfC to \ukC, \dgC, and \ilC to \ghC. 

For \underline{Adaptation Testing}, the pretraining tasks were:

\begin{itemize}
\item 1 Cell: \mgA

\item 4 Cells: \mgA, \nfA, \rsC, \waA

\item 8 Cells: \mgA,  \mfA, \glA, \ocA, \lgA, \mfC, \ddC, \voC

\item 16 Cells: \mgA, \roA, \nfA, \ocA, \itA, \voA, \slA, \lgA, \mfC, \mgC, \meC, \rsC, \voC, \lgC, \lcphxpC, \slC

\item 32 Cells: \meA, \mgA, \mfA, \vzA, \roA, \rsA, \ggA, \nfA, \glA, \tiA, \sgA, \ocA, \ddA, \itA, \voA, \slA, \waA, \lgA, \roC, \mfC, \mgC, \vzC, \rsC, \ggC, \sgC, \itC, \waC, \voC, \tiC, \lgC, \lcphxpC, \slC
\end{itemize}

For \underline{Ablations}, the pretraining tasks were \meA to \ggA. 

\subsection{Individual Figures}


\textbf{Distribution Of Input String Length, Figure \ref{fig:stringlen_distribution}:}
For this, we use the evaluation results from \underline{Limit Testing} and simply make the histogram of the inputs used across all cells during test-time.

\textbf{Outlier Sensitivity In Rank Correlation, Figure \ref{fig:new_adaptation_mseexclusion}:} We used the 8-cell pretrained model from \underline{Adaptation Testing}, and also use the same cells for evaluation. Note that all cells are technically ``out-of-distribution'' for a randomly initialized checkpoint.

We use the same fine-tuning procedure for all model tests. We repeat the fine-tuning procedure 10 times and report the mean for the Spearman rank correlation. We pick a single seed for our Diagonal Fit scatter-plot.

\textbf{Density Capture by RLM, Figure \ref{fig:new_idcell_kdecloud}:} From \underline{Adaptation Testing}, we use the 8-cell out-of-distribution result. We take the predictions as well as samples (128 samples per input), and make a density plot of the metric (MIPS per GCU).

\textbf{MSE Histogram for In and Out-Of-Distribution Adaptation, Figure \ref{fig:new_histograms_all}:} We use the \underline{Limit Testing} setting, with pretrained model checkpoint at 15K steps, which achieved the lowest validation loss within 5K step intervals. Recall from Section \ref{subsec:epistemic_uncertainty} that to obtain the theoretically optimal tabular and null residuals, we grouped all $y$'s into the equivalence classes and took the mean $y$-value of each equivalence class as the theoretically optimal prediction.






\textbf{Pretraining Diversity and Few-Shot Adaptation, Figure \ref{fig:main_finetuning}:} From \underline{Adaptation Testing}, we plot all of the results. For fine-tuning experiments, we use pretraining checkpoint early stopped at 10K steps, motivated by our checkpoint-learning rate analysis in Figure \ref{fig:mse_checkpoint_scan}, and observing validation loss.





\textbf{Prediction Uncertainty and Multi-modality, Figure \ref{fig:sample_uncertainity_msemean} and Figure \ref{fig:bimodal_demo}:} We simply fine-tune the 8-cell pretrained base model from \underline{Adaptation Testing} on in-distribution task \mgA for 512 samples, and pick a seed. 

For Figure \ref{fig:sample_uncertainity_msemean}, we report the scatter plot of prediction standard deviation across its 128 generated samples per input on the Y axis as prediction uncertainty, and squared-error on the X axis. 

For Figure \ref{fig:bimodal_demo}, we identify a set of time-stamps where the ground truth target demonstrates bi-modal distribution, and pick one of these time-stamps. At this exact time-stamp, we have one input and 128 generated regressor predictions. We plot the ground-truth values observed at that time-stamp, as well as the 128 generated values (Sampled Predictions). We then plot the histogram and density plot on the sampled predictions.

\textbf{Scatter and Density Plots, Figure \ref{fig:many_diagonals} and Figure \ref{fig:many_kdeplots}:} We use the \underline{Limit Testing} setting, with pretrained model checkpoint at 15K steps. We fine-tune the model for each task individually, over 512 examples, with a learning rate of $1\times 10^{-4}$ for a single seed.

\textbf{Explained Variance and Negative Log-Likelihood, Figure \ref{fig:percell_finetuned_MSE_result}:}
We use the \underline{Limit Testing} setting, with pretrained model checkpoint 15K steps. To reliably estimate $\text{NLL}_{\text{null}}$, we fine-tune the randomly initialized variant of the \underline{Limit Testing} model on \textit{empty strings} with 1024 examples. $\text{NLL}_{\text{RLM}}$ is also fine-tuned on 1024 examples. Learning rate is $1\times 10^{-4}$, and we report results over 5 seeds.

\textbf{Validation Loss and MSE, Spearman $\rho$, Figure \ref{fig:ckpt_valloss_mse} and Figure \ref{fig:ckpt_valloss_spearmanrho}:} We train a new base model, with 4 layers, 4096 sequence length, on high spread cells (\mgA, \mfA, \lcphxpA, \rsA, \ggA, \nfA, \glA) and save checkpoints at 1K step intervals. We then evaluate every checkpoint on in-distribution cell \mgA over 100 seeds, without any fine-tuning. We report the mean MSE and Spearman-$\rho$ over 100 seeds. 


\textbf{Encoder-Decoder Ablation, Figure \ref{fig:val_loss_encdec}:} We train four models on the \underline{Ablation} setting with the same default model dimensions, but with different encoder and decoder layers. \texttt{(0E4D, 1E3D, 3E1D, 2E2D)} had roughly the same number of parameters \texttt{(62.3M, 60.2M, 58M, 56M)} respectively, and minimum validation loss was observed at step \texttt{(18K, 17K, 17K, 23K)} respectively. 

\textbf{Parameter Size Ablation, Figure \ref{fig:val_loss_param}:}
We train four models, of sizes \texttt{(45.5M, 58.1M, 83.2M, 234.3M)} (2 layers, 4 layers, 8 layers and 32 layers respectively), with default model dimensions on the \underline{Ablation} setting, but with a changed sequence length of 1024. We then measure the minimum validation loss, achieved at respective steps (27K, 16K, 13K, 11K).



\textbf{Sequence-Length Ablation, Figure \ref{fig:val_loss_seq_len}:} We train six models at sequence lengths of \texttt{(4096, 2048, 1024, 768, 512, 256)}, with default model size and dimensions on the \underline{Ablation} setting. We measure the minimum validation loss, achieved at respective steps \texttt{(26K, 23K, 16K, 18K, 13K, 12K)}.

\textbf{Feature Ablation, Figure \ref{fig:val_loss_features}:}
We train three models all of default dimensions but with a changed sequence length of 1024, and measure the minimum validation loss, achieved at step \texttt{(22K, 20K, 23K)} for \texttt{(CR, R, WR)} respectively on the \underline{Ablation} setting.

\textbf{Checkpoint and Learning Rate Ablation, Figure \ref{fig:mse_lr_ckpt_crossover} and Figure \ref{fig:mse_checkpoint_scan}:} We use the default model size but with a sequence length of 4096. This model is trained on \mgA to \glA{} and then fine-tuned and evaluated on the highest-spread out-of-distribution task \roC{}.

For Figure \ref{fig:mse_lr_ckpt_crossover}, we pick the checkpoint at 10K steps and fine-tune the base model on \roC{} at learning rates of $(5\times 10^{-3}, 1\times 10^{-3}, 5\times 10^{-4}, 1\times 10^{-4}, 5\times 10^{-5}, 1\times 10^{-5}, 5\times 10^{-6}, 1\times 10^{-6})$ for \texttt{(0, 4, 8, 16, 32, 64, 128, 256)} examples each. We report the mean of MSE over 10 fine-tuning seeds. 

For the checkpoint ablation in Figure \ref{fig:mse_checkpoint_scan}, we set a fine-tune learning rate of $5\times10^{-5}$ and take checkpoints at every-step, starting from 1K up to 100K. We fine-tune every checkpoint of the base model on \roC{} with validation-loss based early stopping, for \texttt{(0, 4, 8, 16, 32, 64, 128, 256)} fine-tuning examples. Due to the scale of the study, we only use one seed.

